%% file: main.tex
\pdfoutput=1
\documentclass[]{alaya}
\usepackage{makecell}
\usepackage{wrapfig}
\usepackage{tabularx}
\usepackage{textcomp}
\usepackage{stfloats}
\usepackage{url}
\usepackage{verbatim}
\usepackage{titlesec}
\usepackage{tocloft}
\usepackage{adjustbox}
\usepackage{multirow}
\usepackage{pifont}
\usepackage[sc]{mathpazo}
\usepackage{tikz}
\usepackage{comment}
\usepackage{amsmath,amssymb}
\usepackage{colortbl}
\usepackage{natbib}
\usepackage{color}
\usepackage{booktabs} 
\usepackage{hyperref}
\usepackage{graphicx}
\usepackage{subcaption}
\RequirePackage{xspace}
\makeatletter
\DeclareRobustCommand\onedot{\futurelet\@let@token\@onedot}
\def\@onedot{\ifx\@let@token.\else.\null\fi\xspace}
\usepackage[most]{tcolorbox}
\usepackage{array}
\usepackage{siunitx}
\usepackage[table]{xcolor}
\usepackage{caption}
\definecolor{headerpurple}{HTML}{d8d2fc}
\definecolor{rowgray}{gray}{0.95}
\usepackage{CJKutf8}

\makeatother

\definecolor{adptorange}{RGB}{248, 205, 172}
\definecolor{cmpblue}{RGB}{189, 215, 238}

\definecolor{our_red}{RGB}{232,157,160}
\definecolor{our_blue}{RGB}{136,206,230}
\definecolor{our_orange}{RGB}{246,200,168}
\definecolor{our_green}{RGB}{178,211,164}

\definecolor{attn_code0}{RGB}{247,215,200}
\definecolor{attn_code1}{RGB}{238,169,139}
\definecolor{mlp_code0}{RGB}{204,201,221}
\definecolor{mlp_code1}{RGB}{102,95,153}
\definecolor{mygray}{HTML}{f0f0f0}

\definecolor{token_blue}{RGB}{84, 120, 140}

\usepackage{bbding}
\usepackage{fontawesome}
\usepackage{float}

\newlength\savewidth

\newcolumntype{x}[1]{>{\centering\arraybackslash}p{#1pt}}
\newcolumntype{y}[1]{>{\raggedright\arraybackslash}p{#1pt}}
\newcolumntype{z}[1]{>{\raggedleft\arraybackslash}p{#1pt}}

\renewcommand{\paragraph}[1]{\vspace{1.25mm}\noindent\textbf{#1}}

\usepackage{algorithm}
\usepackage{listings}

\definecolor{codeblue}{rgb}{0.25, 0.5, 0.5}
\definecolor{codekw}{rgb}{0.35, 0.35, 0.75}
\lstdefinestyle{Pytorch}{
    language = Python,
    backgroundcolor = \color{white},
    basicstyle = \fontsize{9pt}{8pt}\selectfont\ttfamily\bfseries,
    columns = fullflexible,
    aboveskip=1pt,
    belowskip=1pt,
    breaklines = true,
    captionpos = b,
    commentstyle = \color{codeblue},
    keywordstyle = \color{codekw},
}

\definecolor{green}{HTML}{009000}
\definecolor{red}{HTML}{ea4335}

\usepackage{enumitem}
\definecolor{PanelAccent}{HTML}{D55E00}   

\newcommand{\methodname}{\textsc{Marionette}}

\title{Marionette: Predicting World States, Rendering Geometry, Painting Appearance}
\author[1,2,3,*]{Zian Meng}
\author[1,*]{Zhen Li}
\author[1]{Chuanhao Li}
\author[3]{Qiang Li}
\author[1,2,\dagger]{Kaipeng Zhang}
\affiliation[1]{Alaya Lab}
\affiliation[2]{Shanghai Innovation Institute}
\affiliation[3]{Huazhong University of Science and Technology}
\affiliation{\mbox{$^{*}$Equal contribution}}
\affiliation{\mbox{$^{\dagger}$Corresponding author}}

\abstract{
Interactive game world models typically autoregress visual observations directly in pixel or latent space, forcing structured properties such as pose, geometry, and occlusion to be implicitly maintained by the same generative sequence.
Over long horizons, errors in these latent world properties accumulate, making consistency
and controllability fragile.
We explicitly model the evolving world state, delegate exact geometric computation to a fixed, zero-parameter renderer, and leave the
neural model to synthesize appearance.
We instantiate this idea as \methodname, a world model for interactive games with articulated characters, consisting of three components.
First, a two-stage autoregressive \emph{dynamics model} predicts an explicit and interpretable $276$-dimensional 3D world state comprising multi-entity articulated skeletons, metric root trajectories, and rotations.
Second, a zero-parameter \emph{graphics bridge} converts the predicted state into
pose-control videos, computing world-space geometry and occlusion in closed form.
Third, a control-conditioned video-diffusion \emph{observation model} synthesizes photorealistic RGB observations from the resulting structured controls.
Our experiments establish two properties of \methodname.
First, the predicted world state is directly controllable. Forcing a mismatched action stream changes root-aligned joint error by $31\%$ across $48$
held-out segments, so the articulated dynamics respond to the actions they are given.
Second, long-horizon behaviour is determined in the state, and can be repaired there. Left free,
the two generated characters drift to $21.2$\,m apart (recorded sessions stay near $5$\,m) and a
third of frames show ground penetration. Two rules imposed on the explicit state, a terrain collider and a separation cap, cut penetration by $66\%$ and keep the pair engaged, with no
change to the observation model.
Routing appearance through the predicted state costs no fidelity we can detect, at an FVD of $831$ against $799$ for recorded pose.
Together, these results show that \methodname{} provides an explicit and controllable world representation while preserving high-quality visual generation.
}

\date{\today}
\github{\url{https://alayalab.github.io/Marionette/}}

\begin{document}
\maketitle
\vspace{1.4cm}  

\section{Introduction}
\label{sec:intro}

Interactive game world models generate a playable video stream directly from a stream of user actions, and recent systems reach high visual quality \citep{genie2024,gamengen2024,diamond2024,matrixgame2025,gamegenx2024,oasis2024,genie3_2025,wham2025}.
They do this by autoregressing appearance in pixel or latent space, which leaves everything that must stay consistent, pose, geometry, occlusion, object
identity, and the effect of a control input, to implicit maintenance by the same generative
sequence.
Over a long rollout the model is repeatedly fed its own output, so errors in these implicit properties compound \citep{ross2011dagger,bengio2015scheduled}, and consistency and controllability degrade as the horizon grows.

A video is a high-dimensional observation of a low-dimensional world state.
Learning $p(\text{video})$ directly entangles physics, identity, and control in pixel space.
Positing an explicit world state $s$ makes dynamics and appearance conditionally independent, which splits the problem into two well-posed sub-problems \citep{worldmodels2018,dreamerv32023},
\begin{equation}
p(\text{video}) = \int \underbrace{p(\text{obs}\mid s)}_{\text{observation model}}\;
\underbrace{p(s_{t+1:}\mid s_{\le t}, \text{control})}_{\text{dynamics model}}\; ds .
\label{eq:factorization}
\end{equation}

Where to draw the line between the two should follow what each side does well.
Neural generative models are strong at appearance and perceptual plausibility, and weak at exact bookkeeping over long horizons and at discrete logic \citep{faithfate2023,outofsight2026,mindbench2026,mbench2026}.
Minecraft makes the gap concrete \citep{mineworld2025}.
It includes redstone, a built-in system of logic circuits that behaves like digital wiring.
Whether a redstone lamp turns on is determined by the logical state of that circuit rather
than by the appearance of nearby pixels.
A generator trained to produce the next frame can render one plausible frame while ignoring that state, which is why such logic is better delegated to a deterministic component.
We therefore assign the parts of a world that must be exact, among them geometry, occlusion, and metric motion, to a deterministic renderer, used the way a language-model agent uses a calculator
\citep{react2023,toolformer2023,pal2023}, and leave appearance to the neural model.
Work that must be exact is executed by a tool, not generated.
Appendix~\ref{app:bridge} argues that constraining \emph{what recurs} is more effective than regularizing an unconstrained generator, together with the negative result that shows where it stops.

\begin{figure*}[t]
  \centering
  \includegraphics[width=\linewidth]{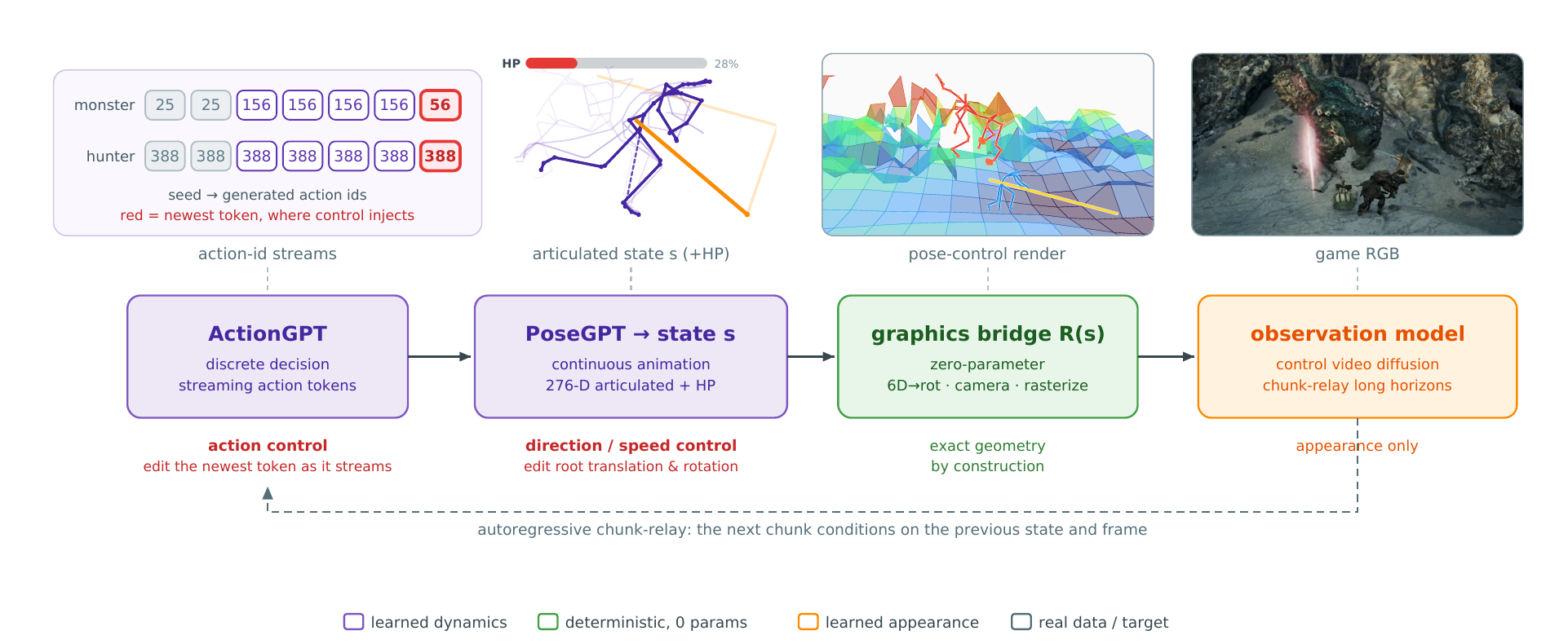}
  \caption{\textbf{\methodname{} overview.} Bottom: the pipeline. ActionGPT makes
discrete per-entity action decisions as a streaming token sequence; PoseGPT turns them into a continuous articulated world state; a zero-parameter graphics bridge renders the state to a pose-control video; a control-conditioned video-diffusion observation model adds appearance. Top: each stage illustrated with recorded data from one moment of a hunter--monster fight. The pose-control render and the game RGB show the same recorded frame through the same camera. Control enters the neural side at two points and nowhere else: editing the newest action token as it streams, and editing the root translation and rotation of the state. Everything the bridge computes from the state is exact by construction; appearance is not, and is the observation model's to supply.}
  \label{fig:overview}
\end{figure*}

\paragraph{\methodname.}
We instantiate this idea as \methodname, a world model for interactive games with articulated characters, with three components (Figure~\ref{fig:overview}).
First, a two-stage autoregressive \emph{dynamics model} predicts an explicit, interpretable world state $s\in\mathbb{R}^{276}$ that encodes two articulated entities, a monster and a player character, as metric root trajectories, root-relative joint positions, and 6D root rotations.
A compact \emph{decision} model (ActionGPT) selects a discrete action per frame and entity,
and a larger \emph{animation} model (PoseGPT) turns the chosen actions into body pose.
Second, a \emph{zero-parameter deterministic bridge} reconstructs metric world-space skeletons from the state and rasterizes them into a pose-control video, computing geometry and occlusion in closed form.
Third, a control-conditioned video-diffusion \emph{observation model} renders that video into photorealistic RGB, relayed across chunks for long horizons.
Because the action is an explicit token, control is applied by overwriting it, which is the same operation a player's button press performs.

Our experiments establish two properties of \methodname.
First, the predicted state is directly controllable. A mismatched action stream changes root-aligned joint error by $31\%$ over $48$ held-out segments, so the articulated dynamics respond to the actions they are given.
Second, long-horizon behaviour is determined in the state. Driving one observation model from recorded state and from predicted state isolates what the dynamics contributes, and rules imposed on the state repair the failures a free rollout produces, without touching
the observation model.
Routing appearance through the predicted state costs no fidelity we can detect, at an FVD of $831$ against $799$ for recorded pose. Both properties are measured in metres, on a world state that a pixel-only model does not expose.

\section{Related Work}
\label{sec:related}

\paragraph{Generative game world models in pixel or latent space.}
A large family of interactive world models autoregresses appearance directly in pixel or latent space, conditioned on user actions, to produce a playable video stream.
The line runs from action-conditional prediction in Atari and GAN-based game simulators \citep{oh2015action,chiappa2017recurrent,gamegan2020,playable2021,promptable2024}, through neural game engines and interactive generators \citep{gamengen2024,diamond2024,oasis2024,mineworld2025,gamegenx2024,wham2025,ivideogpt2024}, to foundation-scale systems targeting open worlds, real-time streaming, and instruction following \citep{genie2024,genie3_2025,matrixgame2025,matrixgame2_2025,thematrix2025,gamecraft2025,gamecraft2_2025,gamefactory2025,astra2026}.
Dedicated surveys now cover this area \citep{igvsurvey2025,sorasurvey2024}, and the action interface is itself widening, through latent actions learned without labels, language interfaces for multi-entity combat, and NPC-directed control \citep{latentactions2026,incantation2026,reactivegwm2026}.
These systems learn rendering, dynamics, and control in one sequence model, and set the
visual standard for interactivity.
But what must stay consistent is held only implicitly in the network's activations, so consistency and controllability are \emph{emergent} rather than guaranteed.
We share the goal of a playable, controllable world and move the exact bookkeeping out of
the generator.
Commercial general-purpose video generators now largely match real footage in open-ended
visual fidelity, but they expose no interface for the per-frame structured control this setting needs, so we treat them as a qualitative reference (Appendix~\ref{app:baseline}).

\paragraph{Separating structure from appearance.}
One line grounds video generation in explicit 3D or in memory, caching generated content, retrieving past context, or conditioning on reconstructed geometry \citep{worldmem2025,spatialmem2025,gen3c2025,contextmem2025,vmem2025,spatia2026,statespace2025,framepack2025}.
A related thread specifies camera trajectories explicitly \citep{motionctrl2024,cameractrl2025,camco2024,cami2v2024}, and recent benchmarks probe how well a generated world preserves off-screen state and responds to actions \citep{outofsight2026,mindbench2026,mbench2026}.
There the 3D structure is a cache or consistency prior derived from pixels that have already been generated, while the forward dynamics stay in the video latent.
A second line predicts structure before rendering it, in driving \citep{mad2026,dreamland2025}, in human video \citep{mosa2025}, and over a persistent semantic-voxel state paired with a learned shader \citep{persist2026}.
Several position papers argue that a generative world model should expose a structured, rule-governed interface and delegate exact simulation to a deterministic component \citep{actionsim2026,igv2025,neurosym2026}.
A concurrent survey organizes the field around the action-state-observation loop of conventional game engines \citep{pixels2states2026}, and mechanistic analyses read video generators through the same state-and-dynamics lens \citep{mechanistic2026}.
These works point in the same direction as ours.
We place the \emph{forward} model in an explicit metric articulated state, and turn that state into appearance with a graphics bridge that has no learnable parameters.
Pose-guided character synthesis \citep{animateanyone2023,champ2024,mimicmotion2024} and the enhancement of engine buffers to photorealistic video \citep{richter2021} establish that such a control signal is enough to drive high-quality video, which our observation model relies on. There the driving signal is an \emph{input}, whereas a world model has to predict it and roll it out.

\paragraph{Controllable articulated motion.}
Character animation now generates articulated motion directly, under text and kinematic constraints.
Diffusion and discrete-token sequence models turn a text description into a motion clip \citep{tevet2023mdm,zhang2024motiondiffuse,zhang2023t2mgpt}.
The control interface has since narrowed from a whole-clip prompt to a chosen joint at a chosen frame \citep{xie2024omnicontrol}.
A parallel line drives characters through a physics simulator, tracking reference clips and reusing learned skill embeddings \citep{peng2018deepmimic,peng2021amp,peng2022ase,luo2023phc}.
Ground and scene conditioning has a long history here, from locomotion controllers that take the local terrain profile as input \citep{holden2017pfnn} to models that place the body against scene geometry \citep{starke2019nsm,hassan2021samp}.
ARDY \citep{ardy2026} streams motion autoregressively with a diffusion denoiser over a hybrid representation that pairs an explicit root with a latent body embedding, and accepts long-horizon constraints such as waypoints, paths, and keyframe poses.
Kimodo \citep{kimodo2026} scales a kinematic motion diffusion model over several hundred hours of professional capture, driving multiple skeleton conventions from text plus end-effector, waypoint, and keyframe constraints.
Turning a control signal into articulated motion is the sub-problem our factorization
isolates, and these models solve it at production-animation quality.
This shows that the dynamics half of Eq.~\ref{eq:factorization} is a well-posed target on its own.
Either model could supply the animation half of our dynamics stage directly, since the
interface between our two stages is a state sequence and nothing more.
ARDY splits an explicit metric root from a learned body representation just as our state
does, because the trajectory must be controlled in world units while the pose need not be.
Our setting adds four things.
Two entities fight each other.
The model predicts the next action itself, and a control input overwrites that prediction at a single token.
Both stages are conditioned on the scanned height field of the shipped game map.
Finally, a deterministic bridge carries the state past the skeleton to rendered video.

\paragraph{Which ``world model'' we mean.}
The term spans distinct goals. One long-standing use is control-oriented, where latent world models learn an abstract state
and predict short-horizon futures in it so that an agent can plan \citep{worldmodels2018,planet2019,dreamerv32023,dreamer4_2025,navwm2025,dinowm2025}, and for that purpose the state serves the policy and need not be exact or readable by a person.
Occupancy world models predict an explicit 3D scene state, chiefly for driving perception \citep{occworld2023,wovogen2023}, and general-purpose video foundation models are increasingly framed as world simulators for physical domains \citep{sora2024,cosmos3_2026,physlaw2025}.
We keep the classical dynamics-and-observation decomposition, but we target a different regime.
Ours is an interactive and persistent game world, where the objective is consistency and controllability.
That regime calls for a state which is both \emph{interpretable} and \emph{render-ready}, so that one explicit state serves forward prediction and photorealistic synthesis alike.

\section{Method}
\label{sec:method}

\subsection{Problem formulation}
\label{sec:method:problem}
We represent the world by an explicit, interpretable state $s_t \in \mathbb{R}^{276}$ that describes the physical configuration of the scene at frame $t$.
For each of the two articulated entities, a monster $M$ and a player character (the hunter) $N$, it holds a per-frame root displacement, joint positions relative to that root, and the root orientation as a continuous 6D rotation, together with a small weapon sub-state:
\[
s = [\,\underbrace{\delta^M_{0:3}}_{\text{root}},\
\underbrace{p^M_{3:162}}_{53\times 3\ \text{joints}},\
\underbrace{\delta^N_{162:165}}_{\text{root}},\
\underbrace{p^N_{165:258}}_{31\times 3\ \text{joints}},\
\underbrace{w_{258:264}}_{\text{weapon}},\
\underbrace{r^M_{264:270}}_{\text{6D rot}},\
\underbrace{r^N_{270:276}}_{\text{6D rot}}\,].
\]
Storing joints relative to the root, and root motion as a per-frame delta, makes the representation invariant to global position and heading.
It also makes the state \emph{render-ready}, since a fixed operator can reconstruct metric
world-space joints from $s$ (\S\ref{sec:method:bridge}).
The control signal $c_t$ carries, per entity, a discrete \emph{action} id from an action
vocabulary, naming what the character is doing (attacking, dodging, walking), and the
character heading; the root displacement and rotation can be overwritten the same way
(Appendix~\ref{app:dynamics}). All of it enters the dynamics model.

Three components then realize Eq.~\ref{eq:factorization}.
A \emph{dynamics model} $p(s_{t+1:}\mid s_{\le t}, c)$ predicts future states (\S\ref{sec:method:dynamics}), a deterministic render operator $R(s)$ produces a pose-control video (\S\ref{sec:method:bridge}), and an \emph{observation model} $p(\text{obs}\mid R(s))$ synthesizes RGB (\S\ref{sec:method:obs}).
The offloading principle fixes where the cut falls. Metric geometry, root integration, joint
kinematics, and camera projection are all computed by the zero-parameter operator $R$, and the neural models are responsible only for plausible dynamics and appearance.

\subsection{Two-stage dynamics: decision then animation}
\label{sec:method:dynamics}
The dynamics model factors the next-state prediction into a \emph{discrete decision} stage and a \emph{continuous animation} stage. This is a small instance of the decoupling principle.
The decision of what to do is discrete and low-dimensional; the body motion that carries it
out is continuous and high-dimensional.

\paragraph{ActionGPT (decision).}
A compact causal transformer ($4$ layers, width $256$, $\sim$2.5M parameters) autoregresses one discrete action token per frame per entity, drawn from that entity's action vocabulary.
It is conditioned on a low-dimensional summary of recent state: the $18$D root/rotation stream, learned action embeddings, animation progress, and the weapon sub-state. It additionally regresses the next root motion and orientation. Because the action is an explicit token, control is applied by overwriting that token. To make a character perform a desired action we override the sampled token with the target action id at that frame, with no retraining and no auxiliary conditioning network. A player's button press would enter the model at exactly this point, which is also where our
scripted control experiments inject their commands; \S\ref{sec:exp:ctrl} shows that the forced
token has causal authority over the resulting pose.

\paragraph{PoseGPT (animation).}
A larger causal transformer ($8$ layers, width $512$, $\sim$25M parameters) maps the chosen action tokens and recent body state to the next-frame $258$D body pose (the two entities' relative joint positions and the weapon). It acts as an animation player, mapping the chosen action to the resulting body configuration. This split keeps the control interface compact and interpretable, while pose synthesis gets
the capacity it needs.

\paragraph{Autoregressive rollout.}
At each step ActionGPT proposes actions and root motion, and an optional control override
replaces the action tokens. PoseGPT animates the body for those actions, the two are assembled
into the next $276$D state, and the state is appended to the context. The loop repeats within
a bounded attention window. Optional terrain conditioning, in the form of an egocentric height patch and per-joint ground clearances encoded by a small MLP, feeds both stages, and the animation stage carries ground-contact and non-penetration objectives so that the predicted motion respects the local ground surface.

\subsection{Deterministic graphics bridge}
\label{sec:method:bridge}
The bridge $R$ turns a state sequence into a pose-control video with \emph{zero learnable parameters}. For each frame it converts the 6D root rotations \citep{zhou2019cont} to rotation matrices by Gram--Schmidt orthonormalization, then rotates each per-frame root delta into the world frame and integrates it by cumulative sum to recover the absolute root trajectory.
It places every relative joint by adding the root-frame-rotated offsets to the root position, which yields metric world-space skeletons for both entities.
It then views the scene with a camera, by default a deterministic follow-camera that frames
the two entities, or a supplied ground-truth view for evaluation, and rasterizes the skeletons into the pose-control frame the observation model consumes.

Every step here is a closed-form geometric operation, so world-space consistency, correct occlusion ordering, and metric scale hold \emph{by construction}.
The rendered geometry can drift only if the predicted \emph{state} drifts, which is what the state-layer metrics measure.

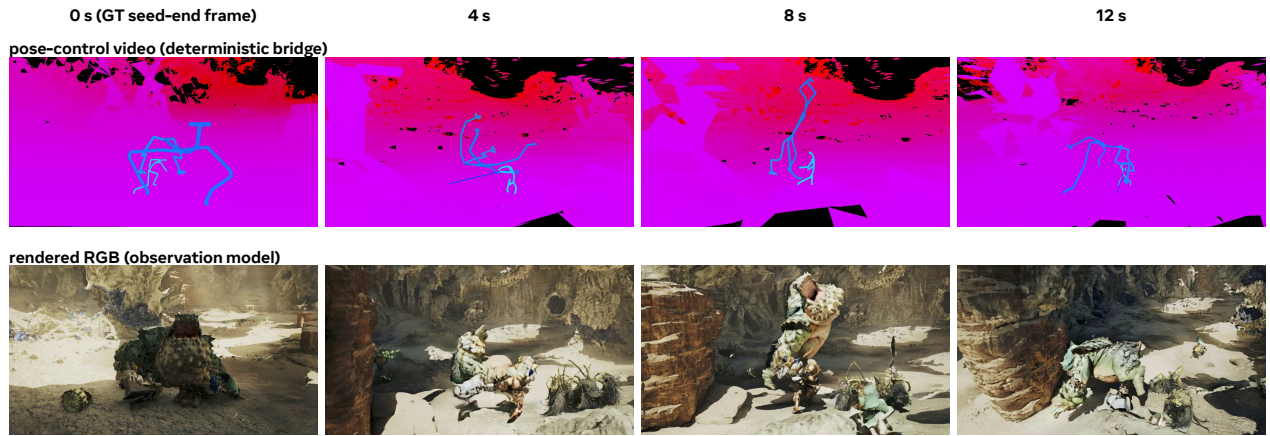
\begin{figure}[t]
\centering
\input{figures/panels/bridge_strip.tex}
\caption{The deterministic bridge on a $12$-second generated rollout. Top:
pose-control frames rasterized from the predicted state (terrain depth encoding with both entities' skeletons). Bottom: the RGB frames the observation model renders from them. The geometry shared by the two rows is computed by the bridge, not generated by a neural model.}
\label{fig:bridge}
\end{figure}

The bridge also carries the scanned terrain, the one part of the world that neither recorded view contains and that both the renderer and
the dynamics model consume.
The state is metric and open to inspection, so the bridge can \emph{enforce} feasibility rather than encourage it, by projecting an infeasible root back onto the ground surface at the frame it appears. The pose-control frame is built as a three-channel geometry buffer, packing height, per-joint identity, and inverse depth into channels the observation model can read back. Appendix~\ref{app:bridge} gives the encoding, the terrain representation, and the effect of both on the training objective and on inference. Figure~\ref{fig:bridge} shows the bridge on a real rollout: the rasterized pose-control frames (top) and the RGB frames the observation model renders from them (bottom), frame-aligned over twelve seconds.

\subsection{Observation model and long-horizon rollout}
\label{sec:method:obs}
The observation model is a control-conditioned video-diffusion model (Wan2.2-Fun-5B) built on a modern DiT video backbone \citep{dit2023,wan2025}, of the family that also underlies recent large video generators \citep{hunyuanvideo2024,cogvideox2025}. It maps the pose-control video from $R$, together with a first-frame image for identity/appearance, to photorealistic RGB. For entities carried in the state it does not need to infer geometry, and the one long-range
quantity it still carries is appearance identity across chunks (\S\ref{sec:limitations}).
Its job is to render the geometry it is given, which is the regime in which pose-guided video synthesis is strong \citep{animateanyone2023,champ2024,mimicmotion2024}. For horizons far beyond a single diffusion window we use a chunk-relay rollout, in the spirit of autoregressive long-video generation \citep{causvid2025,rollingforcing2026}: the model generates a fixed-length chunk and its final frames seed the next one.
Appearance is propagated this way, while the underlying geometry and dynamics stay governed by the explicit state and the deterministic bridge. Because the diffusion model handles appearance only, long-horizon geometric consistency
depends on the state layer, and appearance consistency remains its own
(\S\ref{sec:limitations}).

\section{Experiments}
\label{sec:exp}
Our experiments answer two questions.
\textbf{(Q1)} Does a control input have authority over the generated world? We force an entity's action stream and measure what the body does (\S\ref{sec:exp:ctrl}).
\textbf{(Q2)} What decides how the world behaves as the horizon grows? We hold the observation model fixed and change only the state it is driven with, which is an ablation only a decoupled model admits (\S\ref{sec:exp:ec}).
\S\ref{sec:exp:setup} fixes the data, baseline, and protocol, and \S\ref{sec:exp:duallayer} defines the two evaluation layers.
Additional ablations and full distributions are in the appendix.

\subsection{Setup and evaluation protocol}
\label{sec:exp:setup}
\paragraph{Data and rollout regime.}
Our data is drawn from a commercial action game with articulated player and monster characters. Gameplay corpora with temporally aligned action or state annotations remain scarce \citep{omniworld2026,egocs2026,wildworld2026,pixels2states2026}, and our recordings provide exactly such annotations. The same gameplay recordings yield two synchronized views, an RGB video stream and a per-frame $276$D articulated world state, so that ground-truth state and ground-truth appearance are available for the identical footage. This correspondence is what makes the two-layer evaluation and the bridge ablation (\S\ref{sec:exp:ec}) possible. The dynamics model is trained on the state view: $1{,}395$ segments at $20$\,fps, per-entity action vocabularies of $173$ (monster) and $689$ (hunter), and monster, player, and weapon skeletons of $54$, $32$, and $2$ points.
The observation model is trained on the RGB view under pose-control conditioning at $704\times1280$, with $81$-frame chunks at $30$\,fps and chunk-relay for long horizons. Dataset construction, provenance, the terrain scan, the corpus-wide entity diversity we hold fixed, and the current single-monster scope of the dynamics model are detailed in Appendix~\ref{app:dataset}.

\paragraph{The benchmark.}
We fix one evaluation protocol and use it throughout.
It combines two families of metrics, a \emph{state layer} and an \emph{observation layer} (\S\ref{sec:exp:duallayer}), with a fixed set of seed contexts, rollout horizons, and control-injection scripts.
The observation layer applies to any model that emits RGB; the state layer only to models
that expose an explicit state.

\paragraph{Baseline.}
The quantitative baseline is an \emph{end-to-end pixel-autoregressive} world model trained
on the same RGB footage, conditioned on the same first frame, and rolled out with the
same chunk relay. It receives the scenario as text where our observation model receives the
rendered pose video, and it must carry dynamics, geometry, and appearance jointly in its
latent. We take this to be the closest available comparison for our claim, because it shares
our backbone family and training footage.
What differs is \emph{where} the world state is represented: in an explicit metric state with a deterministic renderer, or in an implicit latent.
Model capacity and RGB training data are shared; our system additionally consumes the state
view through the bridge, which is exactly the design difference under test. The comparison is drawn among systems that share the task as we pose it; a metric computed across mismatched protocols would mainly measure the protocol difference
(scope in Appendix~\ref{app:baseline}).

\paragraph{Metrics and protocol.}
State-layer metrics are computed by reconstructing metric world-space joints from the predicted state (\S\ref{sec:method:bridge}) and comparing to the ground-truth state. Observation-layer metrics are computed on rendered RGB. Because a long rollout of a stochastic world may legitimately diverge from any single reference, we distinguish short-horizon accuracy metrics from long-horizon distributional metrics (\S\ref{sec:exp:duallayer}), and never summarize a long rollout by a single reference-based accuracy scalar (\S\ref{sec:exp:ec}). Full metric definitions are in Appendix~\ref{app:protocol}, and the control-injection
protocol in Appendix~\ref{app:controllability}.

\subsection{Evaluation}
\label{sec:exp:duallayer}

The decoupling lets us measure error \emph{in the world state itself}, in metres, before it is ever rendered. We therefore evaluate at two layers.

\paragraph{State layer (dynamics, absolute units).}
Because the state is a metric articulated skeleton, the predicted dynamics can be scored
directly, in absolute units. Per-frame accuracy against a reference is the wrong target at long horizons, because a plausible world may diverge from any single recorded trajectory, so the state-layer numbers we report are physical-plausibility quantities that need no reference to be meaningful: ground penetration, foot-skate, and how far the two entities have drifted apart. Each is compared with the same measurement on the recorded state of the same windows
(Table~\ref{tab:arms}). Where a control input is involved, RA-MPJPE against the recorded pose is the measure, compared
between control conditions (Table~\ref{tab:controllability}). 

\paragraph{Observation layer (rendering, shared ground).}
In the observation space any model that emits RGB can be compared. We report FVD \citep{fvd2018} for perceptual and temporal quality, both over short clips and as a function of the chunk index over long rollouts. Against the pixel-autoregressive baseline our observation model reaches an FVD of $831$ to its $975$ (Table~\ref{tab:duallayer}), with clip-bootstrap intervals that overlap, so routing every frame through an explicit state and a zero-parameter renderer costs no appearance quality this measurement can detect.
FVD is the only automated observation-layer score we report; reference-based per-frame scores
are ill-posed for stochastic long rollouts, and broad perceptual suites \citep{vbench2023}
target properties that do not bear on our claims.
Whether the intended action happened is evaluated on the state layer, where it can be
measured directly.

\begin{table}[t]
\centering
\caption{Observation layer, against the end-to-end pixel-autoregressive baseline. Twelve $16$-second rollouts per model against real footage; FVD$_{16}$ is over $16$-frame
clips, eight per rollout ($128$ real / $96$ generated),
every pair started from the recorded first frame of its own held-out window, with the bridge
camera initialized from the recorded view; our rollouts use the shipped configuration (terrain
collider on). The final-chunk row uses the last $81$ frames. Bracketed ranges are clip-level bootstrap percentile intervals \citep{efron1979bootstrap};
resampling duplicates clips and inflates a Fr\'echet distance, which is why they sit above the
point estimates. The bias is common to both columns, and only the overlap is read. State-layer
behaviour is in Table~\ref{tab:arms} and Figure~\ref{fig:ablation}, where each number is
read against the recorded state; controllability is in Table~\ref{tab:controllability}.}
\label{tab:duallayer}
\begin{tabular}{lcc}
\toprule
Metric & \methodname{} & Pixel-AR baseline \\
\midrule
FVD$_{16}$ $\downarrow$                        & $831$ $[963,1209]$  & $975$ $[1101,1346]$ \\
FVD, final chunk of long rollout $\downarrow$  & $1013$              & $1198$ \\
\bottomrule
\end{tabular}
\end{table}

\subsection{Q1: does a control input have authority?}
\label{sec:exp:ctrl}
\paragraph{Claim under test.} Our control interface is a discrete action token on the state
stream. The claim is that writing that token changes what the body does.

\paragraph{Design.} We roll out $48$ held-out segments from a common seed context under
three action streams for the hunter. Under \textsc{Free} the model samples its own actions;
under \textsc{Force-GT} we overwrite them with the recorded ids; under \textsc{Force-Shuf}
with a temporally shuffled, and therefore mismatched, copy of those ids. Segment, seed and horizon are
fixed and decoding is deterministic, so the injected stream is the only difference between
conditions. We score RA-MPJPE against the recorded pose with each entity's root subtracted,
so it measures the articulated pose, not the distance travelled.
The baseline exposes no interface for these commands, so the probe has no baseline column;
Appendix~\ref{app:baseline} shows both systems' response to one scripted schedule.

\paragraph{Result.} A mismatched stream degrades the pose by $31\%$, from $0.272$ to
$0.357$\,m, with $\textsc{Force-GT}<\textsc{Force-Shuf}$ on $33$ of the $48$ segments.
Forcing the correct stream matches the free-run level, $0.272$ against $0.281$.

\paragraph{What follows.} The forced token changes the generated pose rather than being ignored, so it has causal
authority over the state. Classes whose poses differ most across ids inflate most, $2.3\times$ stationary against
$1.3\times$ locomotion (Table~\ref{tab:controllability}), the signature of the wrong action
being performed. The second number shows how large that effect is in practice. Forcing the correct token
helps only where the free rollout would have chosen differently, and the small gap means it
seldom does. This establishes the claim in the state space; the full protocol is in
Appendix~\ref{app:controllability}.

\subsection{Q2: what decides long-horizon behaviour?}
\label{sec:exp:ec}
\paragraph{Claim under test.} Two parts. That long-horizon behaviour is determined in the state, and that an explicit
state can be repaired by a rule imposed on it.

\paragraph{Design.} We hold the observation model fixed and change only the state that
drives it, over the same twelve held-out windows, with checkpoint, per-window seeds, and bridge
identical. We evaluate four driving states, the recorded one and three generated ones, differing by
which rule is applied between the dynamics model and the bridge. We score the state layer by ground penetration, foot-skate, and how far the two characters
drift apart, and the observation layer by FVD. The recorded state provides the reference level for each state-layer number.

\paragraph{Result: the failure.} Over the rollout the generated hunter and monster drift apart, from $4.9$\,m at the first frame
to $21.2$\,m at the last. The camera frames both characters, so it pulls back as they separate:
between the first and the last rendered chunk, each body's share of the pose-control frame
falls from $1.81\%$ to $0.57\%$; under recorded state that share holds at $1.55\%$ to $1.60\%$. The generated root
stays on scanned ground in $100\%$ of frames, so the two characters drift apart inside the map.

\paragraph{The rule.} When a frame's motion would carry the two characters more than
$6$\,m apart, we cancel the outward part of that motion and split the correction evenly
between them. It takes the same form as the terrain wall-block (Appendix~\ref{app:bridge}),
its threshold is a fixed constant, and it reads no recorded trajectory, so it adds no
information about the future it is scored against.

\begin{table}[t]
\centering
\caption{The effect of each rule. Twelve held-out windows, identical checkpoint, bridge
commit and seeds; only the rule applied to the state differs. Recorded state is the same
measurement on the recorded trajectories of those windows, and is a reference level, not a
competing method. \textbf{CFR}, collision-frame ratio: frames with any of $11$ key joints
below terrain at a $0.15$\,m margin. \textbf{MMP}, mean maximum penetration depth.
\textbf{Skate}, contact-gated foot slide. \textbf{Sep}, monster--hunter distance at the final frame; the separation cap bounds it at $6$\,m
by construction.}
\label{tab:arms}
\begin{tabular}{lcccc}
\toprule
Driving state & CFR $\downarrow$ & MMP (m) $\downarrow$ & Skate (m/s) $\downarrow$ & Sep (m) \\
\midrule
Recorded \emph{(reference)}        & \emph{0.082} & \emph{0.023} & \emph{1.13} & \emph{4.8} \\
\midrule
Generated, no rule                 & $0.337$ & $0.444$ & $1.14$ & $21.2$ \\
${}$+ collider                     & $0.114$ & $0.157$ & $1.14$ & $21.2$ \\
${}$+ collider + separation cap    & $0.114$ & $0.155$ & $1.22$ & $5.1$ \\
\bottomrule
\end{tabular}
\end{table}

\begin{figure}[t]
\centering
\includegraphics[width=0.82\linewidth]{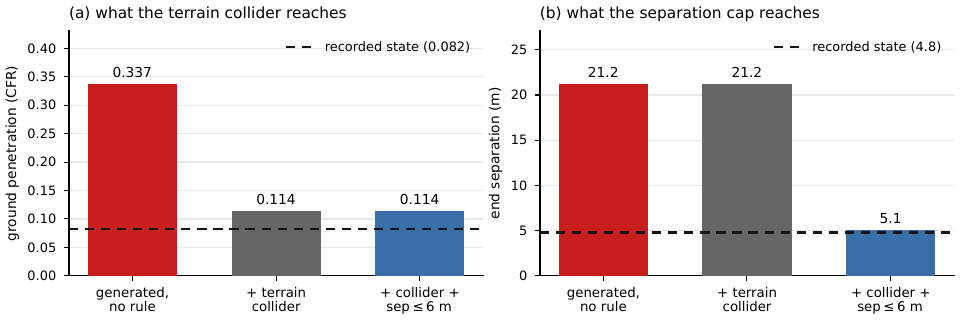}
\caption{Each rule affects only its target quantity. The terrain collider reduces penetration without
changing separation, and the separation cap does the opposite. Dashed line: the same measurement on recorded state
over the same windows.}
\label{fig:ablation}
\end{figure}

\paragraph{Result: the effect of each rule.} The terrain collider cuts the penetration
collision-frame ratio from $0.337$ to $0.114$, a drop of $66\%$, and leaves separation at
$21.2$\,m. The separation cap holds the pair at $5.1$\,m, leaves penetration at $0.114$, and adds $7\%$
of foot-skate, since cancelled motion makes a walking foot slide.
Recorded state scores $0.082$ and $4.8$\,m on the two measures.

\paragraph{What follows.} Each rule governs one degree of freedom and leaves the other
where it was. Together these results support both halves of the claim. The long-horizon failure is a property of the state, and rules imposed on the state repair
it without touching the observation model. Because the state is explicit, each rule is tied to a named degree of freedom and its effect
is checked in metres.

On the observation layer all four states land in a narrow aggregate FVD band, $747$ (with the
cap) to $831$ (collider alone), and none collapses over the horizon, with per-chunk FVD rising by at most a fifth. Rendered differences between them are best judged by eye, and the rollouts are
on the project page.

Substituting recorded state for generated state in the same observation model is a
decomposition available only to a model that carries an explicit intermediate state, since
a pixel-only system has no correct state to substitute. Here it gives $799$ against
$831$ and separates no further at this sample size, which bounds the share of the
observation-layer distance the dynamics can account for.

On the state layer, root divergence from the recorded reference grows steadily with horizon.
That is the expected behaviour of a stochastic world, and the state reports it in metres.
The state layer reports whether a commanded action was executed and how far the state has
drifted; a per-chunk appearance curve reports neither, so the protocol carries both.
Both curves are given in Appendix~\ref{app:protocol}, Figure~\ref{fig:longhorizon}.

\section{Limitations}
\label{sec:limitations}
The limitations below stem from one design decision, the commitment to an explicit,
render-ready state.
Each item below is paired with the change that would address it.

\paragraph{Appearance conditioning.}
Appearance shifts as the horizon grows, while the geometry underneath it stays exactly what
the state prescribes, and the factorization says why.
The pose-control video fixes geometry, depth, and articulation, and the bridge makes those exact, but it says nothing about appearance, so every appearance decision falls to the observation model's own memory, which is propagated
only through the chunk-relay seed and decays as an unconstrained autoregressive quantity should (\S\ref{sec:exp:ec}).
The fix leaves the bridge untouched. Reference images of each entity, persistent and
re-supplied at every chunk, would give the observation model a persistent appearance
reference, playing the role the explicit state plays for the dynamics; our model already
accepts the weakest version of this, a single first-frame reference.

\paragraph{Coverage of the recorded state.}
Our recordings contain entities that appear in the RGB view but have no recorded joints,
notably AI companions and small monsters, so the observation model is trained on pairs in
which part of the target is unexplained by the condition.
Such pairs teach it to supply those entities itself, which at inference appears as content with no state behind it.
This is a property of the capture and is fixable at the source, since the engine already tracks these entities and recording their skeletons brings them into the state and into the render.

\paragraph{Ground-truth and generated pose.}
The observation model is trained on pose-control video rendered from \emph{ground-truth} state, because that is the only pose video for which a paired real RGB frame exists, and at inference it is driven by \emph{generated} state.
The two differ in distribution, since generated pose carries the dynamics model's characteristic errors, so we rely on generalization across that shift, and the bridge ablation of \S\ref{sec:exp:ec}
bounds its observable effect.

Two remedies follow from the argument of this paper.
The first constrains the intermediate representation. The pose video is produced by a
rule-governed operator over a state we can inspect, so out-of-distribution inputs can be projected back into the training distribution before the observation model sees them.
The terrain collider and the separation cap each do this for one failure (\S\ref{sec:exp:ec}), and a fuller set of feasibility rules over rig limits, contact, and joint ranges would do it in general.
The second reduces the shift from the training side, perturbing the ground-truth state when rendering training targets so the observation model learns the neighbourhood of poses it will be asked to render.
The two compose, and neither requires new data.

\section{Conclusion}
\label{sec:conclusion}
We argued that an interactive game world model should fit the \emph{world state} rather than the pixel distribution, delegating the parts that must be exact to a fixed renderer and leaving appearance to the
neural model.
Instantiated for articulated multi-character games, this makes geometry and occlusion exact given the state, and turns control into an overwrite of a single token that moves the generated pose by $31\%$ when the token is wrong.
It costs no appearance fidelity our measurement can detect, at an FVD of $831$ against $799$ for predicted against recorded pose.
The design rests on a bet about long rollouts, that a compounding-error problem is better
addressed by constraining what recurs than by training an unconstrained generator to stay
put.
On that view, an intermediate representation is valuable when rules can be enforced on it,
not when it is compact.
Our own negative result marks the limit of that bet, since a constraint expressed as a differentiable penalty on the body was traded away by the
optimizer, while the same constraint imposed as an input feature and as a post-hoc
projection held.

\subsubsection*{Reproducibility statement}
The gameplay corpus is already public. The recordings, the per-frame action labels, and the
explicit per-entity state are our own previously released \textsc{WildWorld} dataset \citep{wildworld2026}, from which we derive the two synchronized views used here (Appendix~\ref{app:dataset}). The scanned terrain height fields that the bridge consumes (Appendix~\ref{app:bridge}) are new to this work, and we release them alongside our code. We will release the training and evaluation code for the two-stage dynamics model, the deterministic bridge, and the control-conditioned observation model, together with the evaluation protocol (metric implementations, seed contexts, horizons, and control-injection scripts), the dataset manifest and preprocessing, and training configurations for all models and baselines. In this paper the control interface is exercised by scripted injection into the action and root channels (Appendix~\ref{app:dynamics}, Appendix~\ref{app:controllability}) and by offline rollouts, and that is exactly what the released code reproduces. Because the results concern motion over time, the rollouts, the control demonstrations, and the baseline comparisons are also presented as video on the project page, \url{https://alayalab.github.io/Marionette/}, which is the intended way to inspect them. The state layout, model hyperparameters, and rollout procedure are specified in \S\ref{sec:method}. The evaluation protocol and baselines are specified in \S\ref{sec:exp}.

\appendix
\section*{Appendix}
\addcontentsline{toc}{section}{Appendix}

\section{Dataset and Data Integrity}
\label{app:dataset}

\paragraph{Source and the two synchronized views.}
All data derives from recordings of a single commercial action game with articulated player (hunter) and monster characters.
It is drawn from \textsc{WildWorld} \citep{wildworld2026}, the corpus we released for exactly this purpose. The three quantities a world model needs, actions, state, and observations, are produced at
three different points of a running game engine, and WildWorld records each at its source. The engine consumes player \emph{actions}, maintains and updates the \emph{world state}, and renders the state into \emph{observations}. Recording at the engine level is what makes the labels exact, since an action id is the animation the engine actually played instead of a label a vision model guessed, and it is why a state-space dynamics model is feasible here at all. Each recording is processed into two \emph{synchronized} views of the same footage, an RGB video stream and a per-frame $276$-dimensional articulated world state extracted from the game's own pose data. Because both views come from the identical frames, we have ground-truth appearance and ground-truth state in correspondence, which is what enables the two-layer evaluation (\S\ref{sec:exp:duallayer}) and the bridge ablation (\S\ref{sec:exp:ec}). The observation model's pose-control input is produced by rendering the state view through the deterministic bridge (\S\ref{sec:method:bridge}), so training and evaluation use exactly the same state$\to$pose operator.

\paragraph{State (dynamics) corpus.}
The dynamics model is trained on the state view: $1{,}395$ motion segments sampled at $20$\,fps, $\approx$$1.6{\times}10^{7}$ frames in total.
Per-entity discrete action vocabularies hold $173$ entries (monster) and $689$ (hunter), over articulated skeletons of $54$ (monster) and $32$ (hunter) joints plus $2$ weapon points;
each root is carried as a separate displacement channel, leaving $53$ and $31$ root-relative
positions in the state. The state concatenation is given in Table~\ref{tab:state}. Action ids are derived from the game's animation-bank and motion identifiers, which gives a frame-accurate, zero-offset action label.

\paragraph{Scope: single-monster dynamics, multi-monster appearance.}
The dynamics model in this paper is trained on gameplay against a \emph{single} monster type, the most-represented one, whose action vocabulary of $173$ is that monster's action bank.
The multi-monster action space is strongly heterogeneous, with near-disjoint per-monster vocabularies, and single-monster training is the most direct route to a working two-stage model. The RGB observation model, by contrast, is trained across $27$ monster identities and multiple stages and weapon types, since appearance transfers far more readily across monsters than articulated dynamics do. Consequently our end-to-end interactive results are single-monster on the dynamics side. Scaling the dynamics model to the full multi-monster vocabulary is future work (\S\ref{sec:limitations}).

\paragraph{Corpus richness, and the slice we study.}
The source corpus is broader than the subset we use, and we narrow it deliberately. \textsc{WildWorld} spans $29$ monster species, four player characters, four weapon types,
five stages, and party encounters of up to four hunters against one or two monsters
\citep{wildworld2026}. Our own RGB training manifest covers $27$ of those monster species, and spans $5$ weapon
types, $6$ hunter appearances, and $5$ stages, including recordings made after the release. The monster species matter most here, because they are not skin variations of one rig.
Joint counts, bone lengths, proportions, and locomotion modes differ from species to species, so a skeleton, an action vocabulary, and a learned animation prior fitted to one transfer to another only in the loosest sense. For the experiments in this paper we therefore hold most of that variation fixed and study one-on-one combat with a fixed weapon type, so that the state layout, the action vocabulary, and every metric mean the same thing across
all of our ablations. Extending the study is a matter of data rather than method.
Another weapon or another monster is a different slice of the same corpus, re-fitted with the same two-stage recipe, and the bridge needs only that species' rig because it is driven by the state layout and has no learned weights. Widening to a co-operative party is the one case that also touches the layout, since each additional hunter adds its own root, joints, and action stream to the state vector and to both stages' inputs. That is mechanical, but it requires retraining both stages.

\paragraph{Terrain.}
One component of the state that the two views do not carry is the ground itself. A pose sequence says where a body is without saying what it is standing on, and neither an RGB frame nor a joint list tells the dynamics model that a slope rises ahead. We therefore record a third product alongside the two views: a scanned height field of each stage, accumulated from the geometry the engine reports beneath the characters as they traverse it. This is our own addition, produced for this work and absent from the released corpus. The representation, how it enters both the bridge and the dynamics model, and what it does to the losses and to inference are given in Appendix~\ref{app:bridge}.

\paragraph{Hours.}
We quote data volume under one precise definition, the duration of \emph{decodable source footage at native frame rate}, measured per file when the training manifest is built. Under this definition the RGB view used to train the observation model comprises $4{,}008$ unique clips totalling $673.8$ hours, and the state view used to train the dynamics model comprises $1{,}395$ segments totalling $16.4$M frames at $20$\,fps ($227.6$ hours). Hours are counted from unique source footage. Multiplying a window count by a window length would double-count overlapping footage and overstate the corpus size.

\section{World-State Representation}
\label{app:state}
\begin{table}[h]
\centering
\caption{The $276$-dimensional world state $s$. Monster $M$ and hunter $N$ are each
stored as a per-frame root displacement (local), root-relative joint positions, and a 6D root rotation; a small weapon sub-state is stored relative to the hunter. All positions are metric (metres); the world frame is $Y$-up, right-handed.}
\label{tab:state}
\begin{tabular}{lll}
\toprule
Indices & Component & Dimension \\
\midrule
$[0{:}3]$     & Monster root displacement (local) & $3$ \\
$[3{:}162]$   & Monster root-relative joint positions & $159$ \\
$[162{:}165]$ & Hunter root displacement (local) & $3$ \\
$[165{:}258]$ & Hunter root-relative joint positions & $93$ \\
$[258{:}264]$ & Weapon points (hunter-relative) & $6$ \\
$[264{:}270]$ & Monster root rotation (6D) & $6$ \\
$[270{:}276]$ & Hunter root rotation (6D) & $6$ \\
\midrule
& \textbf{Total} & $\mathbf{276}$ \\
\bottomrule
\end{tabular}
\end{table}
Joint positions are stored relative to each entity's root and the root motion as a per-frame local displacement, so the representation is invariant to global position and heading. Absolute world positions are recovered only through the deterministic bridge (Appendix~\ref{app:bridge}). The 6D rotation parameterization stores the first two columns of the root rotation matrix, following the continuous rotation representation of Zhou et al.~\citep{zhou2019cont}.

\section{The Deterministic Graphics Bridge}
\label{app:bridge}
The bridge $R$ maps a state sequence $\{s_t\}$ to metric world-space skeletons and then to a rendered pose-control video, with no learnable parameters. For each entity and frame $t$:
\begin{enumerate}[leftmargin=1.4em,topsep=2pt,itemsep=1pt]
\item \textbf{Rotation.} Convert the 6D root rotation $r_t$ to a matrix $\mathbf{R}_t\in SO(3)$ by Gram--Schmidt orthonormalization of its two 3-vectors ($\mathbf{b}_1=\hat r_t^{(1)}$, $\mathbf{b}_2=\widehat{r_t^{(2)}-(\mathbf{b}_1\!\cdot\!r_t^{(2)})\mathbf{b}_1}$, $\mathbf{b}_3=\mathbf{b}_1\times\mathbf{b}_2$).
\item \textbf{Root integration.} Rotate the local root displacement into the world frame, $\Delta_t^{\text{world}}=\mathbf{R}_t\,\delta_t$, and integrate by cumulative sum, $\mathbf{x}^{\text{root}}_t=\sum_{\tau\le t}\Delta_\tau^{\text{world}}$.
\item \textbf{Joint placement.} Place each root-relative joint $p_{t,k}$ by $\mathbf{x}_{t,k}=\mathbf{x}^{\text{root}}_t+\mathbf{R}_t\,p_{t,k}$, giving metric world-space joints for the monster and hunter skeletons (and the weapon points).
\item \textbf{Terrain.} Bind the stage's scanned height field and tessellate it into a triangle mesh in the same world frame, so that the ground the characters stand on is drawn from measured geometry.
\item \textbf{Camera.} View the scene with a deterministic follow-camera (a fixed geometric function of both
entities' roots, framing the pair) by default, or with a supplied ground-truth camera for
evaluation.
\item \textbf{Projection and rasterization.} Project joints to image space and rasterize the skeletons (bones and joints, with a fixed palette and depth ordering) into the pose-control frame consumed by the observation model.
\end{enumerate}
Every step is a closed-form geometric operation, so world-space consistency, metric scale, and occlusion ordering hold by construction. The rendered geometry can drift only if the predicted \emph{state} drifts, which the state-layer metrics measure directly. The same operator is used to build the pose-control training targets and to render rollouts, so there is no train/test mismatch in the bridge.

\paragraph{What the pose-control frame encodes.}
The rasterized frame carries more than a drawing of a skeleton. It is a three-channel geometry buffer, chosen so that every quantity the observation model needs is recoverable from a colour and nothing is left to be inferred, and so that the ordinary video codec in the training pipeline transports it without destroying the encoding. At $704\times1280$, each fragment is written as
\begin{align*}
R &= \text{height:}\quad 0.5 + 0.5\,\hat h \ \text{ for terrain},\qquad 0.2\,\hat h \ \text{ for a bone},\\
G &= \text{a categorical identity, } \mathrm{id}/255,\\
B &= \text{inverse depth, } 1 - (z - z_{\text{near}})/(z_{\text{far}} - z_{\text{near}}),
\end{align*}
where $\hat h$ is the world $Y$ coordinate of the fragment normalized over the stage's measured height range and $z$ is the metric distance in front of the camera, so near geometry is bright. Three details make this work in practice. First, terrain and bone occupy \emph{disjoint} $R$ ranges. With a single shared range, ground and a limb at the same height receive the same colour and the two collapse into one hue. Splitting them guarantees a separation of at least $0.3$ in $R$, and the mapping stays invertible, since the $G$ channel says which branch to undo. Second, $G$ resolves all the way down to individual joints, with ids drawn from reserved bands: $0$ for terrain, $60$--$99$ for weapon segments, $105$--$194$ for the monster, $195$--$254$ for the hunter.
Each child joint holds a fixed id across the whole rollout. The observation model therefore sees a stable label for ``this limb'' and never has to re-establish correspondence from shape, and the gaps between bands absorb the residual quantization noise of the $10$-bit encoding. Terrain sits at $G\!\approx\!0$ so the ground is chromatically opposite the characters and cannot swallow them. Third, occlusion is resolved by the depth buffer.
Terrain fragments closer to the camera than a few metres ($3$\,m by default) are discarded per frame, so a wall or boulder the follow-camera passes through cannot occlude the character it is meant to show.
Bones are never culled. The frame is rendered into a half-float buffer and encoded at $10$ bits per channel with no chroma subsampling, since $R$ and $B$ are measurements and quantizing them to $8$ bits discards geometric precision that the encoding is there to carry.

\paragraph{Terrain representation.}
Ground is the one part of the world that the two recorded views do not contain, and it cannot be dropped, because an articulated character is only physically meaningful relative to the surface it stands on. We record it separately as a scanned height field, accumulated from the geometry the engine reports beneath the characters as they traverse a stage, and store it as a sparse set of $100\,\text{m}\times100\,\text{m}$ chunks, each a $100\times100$ grid of $1$\,m cells. Each cell holds a short list of \emph{layers} in place of a single height, each layer carrying a height, a surface normal, a hit count, and a validity flag. The layers are what make the representation usable in a real game map.
A single height per cell cannot express a bridge, a ledge, or the floor of a building with terrain above it, and a model querying such a grid under an overhang gets the roof instead of the floor. A query therefore takes a reference height as well as a position and returns the valid layer nearest it, which is the surface the body is actually standing on. Two more operations make the field smooth enough to condition on and to clamp against. Layers within $0.5$\,m of each other in the same cell are merged, so scanning noise does not present as a staircase of near-duplicate surfaces. An unobserved cell is filled by dilation when at least three of its eight neighbours agree to within $2$\,m, so gaps the characters never walked over do not read as holes in the floor. Finally, the height query bilinearly interpolates the four surrounding cell centres. 
The per-cell lookup stepped by up to $20$\,cm at $1$\,m cell boundaries, and because that height clamps the character at inference, the steps appeared in the output as the body visibly snapping upward as it walked.

\begin{figure}[t]
\centering
\includegraphics[width=\linewidth]{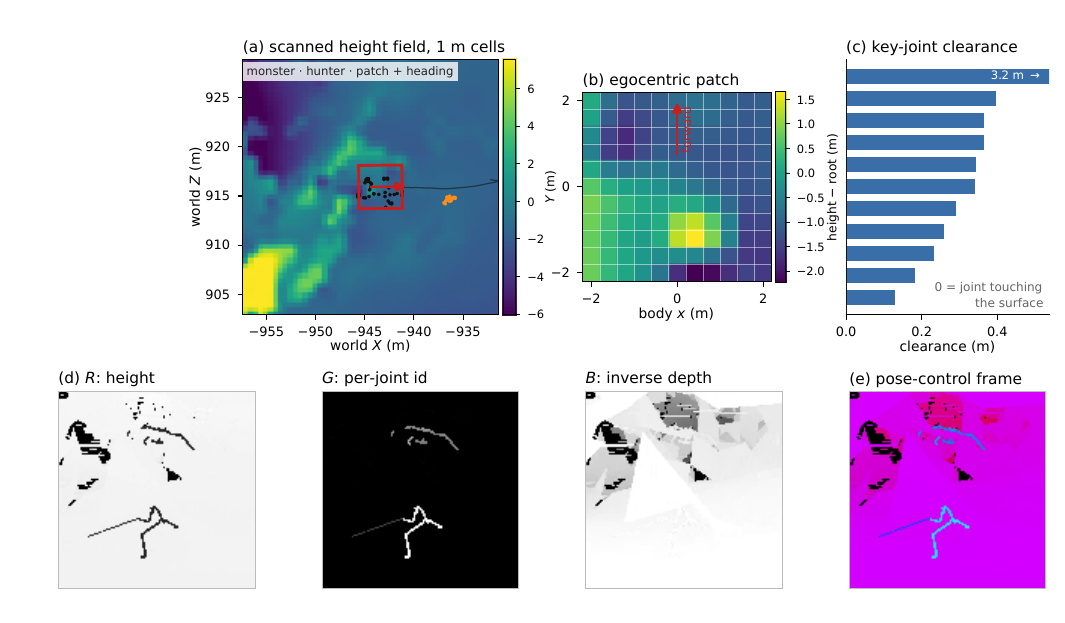}
\caption{Terrain, from scan to conditioning to render. All five panels are one
frame of one real generated rollout. \textbf{(a)}~The scanned height field around the monster, one cell per metre, coloured by world height. The square is the footprint of the region the dynamics model is given and the arrow is the body heading that orients it. \textbf{(b)}~The egocentric patch actually fed to both stages: $11\times11$ cells of $0.4$\,m, heights taken relative to the root, rotated into the body frame and biased $1$\,m forward, so that the same slope yields the same feature at any position and heading. \textbf{(c)}~The second conditioning feature: signed ground clearance of the $11$ key joints, which tells the model which parts of the body are in contact (one raised claw is clipped at the axis limit). \textbf{(d)}~The three channels the bridge rasterizes, shown separately and contrast-windowed: height, in which terrain occupies the upper range and bones a disjoint lower one, so bones read as dark against light ground; per-joint categorical identity, on which terrain is zero; and inverse depth. Black regions are ground the scan never covered, where the render falls through to background. \textbf{(e)}~The composite pose-control frame the observation model receives. Every step from (a) to (e) is a fixed geometric operation with no learned parameters.}
\label{fig:terrain}
\end{figure}

\paragraph{How terrain enters the dynamics model.}
The bridge draws the terrain mesh, but drawing it is not enough, because a model with no access to the ground will predict motion that intersects it. Both stages therefore consume the same two terrain features, re-derived per frame from the body's own root and heading (Figure~\ref{fig:terrain}). In the configuration we report, both are built for the monster entity, whose ground interaction dominates the scene. The construction is per-entity and applies unchanged to the hunter. The first is an \emph{egocentric height patch}: an $11\times11$ grid of $0.4$\,m cells, $4.4$\,m across, centred $1$\,m ahead of the root along the body's forward axis and rotated with its heading.
It is sampled from the height field and stored \emph{relative to the root height}. Making it root-relative and yaw-aligned is what makes it a reusable feature, since the same rising slope produces the same patch anywhere on the map and at any heading, exactly as the joints are stored root-relative for the same reason (Appendix~\ref{app:state}). The second is a vector of \emph{clearances}, the signed vertical distance from each of $11$ key joints of the monster skeleton to the surface beneath it, which tells the model which parts of the body are currently in contact. Both are encoded by a small MLP and concatenated to the stage inputs ($32$-dimensional in ActionGPT, $64$ in PoseGPT). ActionGPT sees them because the decision of where to move next depends on what is walkable. PoseGPT sees them because the pose that realizes a step depends on where the ground is.

\paragraph{Terrain in the training objective, and a negative result.}
Terrain also supplies two auxiliary losses. Neither needs the height field at training time.
For every frame we precompute the terrain height under each key joint, expressed relative to the root, together with the world-up row of the body rotation, which is enough to turn a predicted root-relative joint into a predicted world height. A one-sided hinge $\mathcal{L}_{\text{pen}} = \mathrm{ReLU}\!\left(H_j + \epsilon - y_j\right)^2$ penalizes a joint predicted below the surface, and $\mathcal{L}_{\text{contact}} = \mathbf{1}[\text{contact}]\,(y_j - H_j)^2$ pins joints that the ground truth marks as touching the ground, contact being defined as clearance within $0.25$\,m. A third term acts on the root alone and opposes the cumulative downward drift that an autoregressive root integrator otherwise develops.

The final weighting was set by a failure mode we did not anticipate. With the penetration and contact terms strong (weights $2.0$ and $0.5$), the model did reduce penetration, and it did so by \emph{stretching the skeleton}. Lifting a foot out of the ground is cheaper for gradient descent if the leg simply gets longer, and bone-length error on the monster rose from the rigid ${\sim}3\%$ of the terrain-free baseline to ${\sim}13\%$. The constraint was satisfied only by deforming the rig. The configuration we ship keeps the patch as an \emph{input} and demotes the hinge to a small regularizer (weight $0.1$), drops the contact term entirely, and keeps only the root-level anti-drift term, which cannot deform the body because it does not touch the joints. This mirrors the offloading argument. A rule stabilizes only when it is enforced where the
optimizer cannot trade against it, and a differentiable penalty on a soft, high-dimensional
output is no such place.

\paragraph{Terrain at inference.}
Two things happen at rollout time that cannot happen at training time. First, the patch and the clearances are re-sampled at every step from the model's \emph{own} predicted root and rotation, no longer from a recorded trajectory, so the conditioning stays consistent with the rollout as it drifts and keeps describing the ground the character is actually over. Second, in the shipped configuration the generated root is projected back onto the feasible
set by a collider that runs after the network and before the render; \S\ref{sec:exp:ec}
disables it only as an ablation. It is a simple, fixed rule with no learned parts. A frame's horizontal motion is cancelled if it would step up more than $1.5$\,m onto what is effectively a wall, unless the root is rising fast enough to be a jump. The whole body is then shifted vertically so its lowest joint rests on the surface, with the correction rate-limited to $4$\,cm per frame so that a residual layer switch cannot teleport the character. No part of the collider is learned, and it cannot deform the body, because the same rigid
shift is applied to every joint and to the weapon. The effect is that a dynamics error which would otherwise compound, such as a root sinking into a hillside and then conditioning the next step on an impossible configuration, is removed at the frame it appears.

\paragraph{Why a constrained intermediate representation stabilizes a rollout.}
It is well understood why long autoregressive rollouts degrade. A model trained on recorded frames is fed its own outputs at rollout time, so the input distribution it sees is one it was never trained on.
Each small error moves the next input further out of distribution, and the error compounds \citep{ross2011dagger,bengio2015scheduled}. Pixel-space video generators inherit this directly, and much recent work attacks it inside the generator, through longer or packed context, drift-prevention objectives, and modified forcing schedules \citep{framepack2025,causvid2025,rollingforcing2026,diffusionforcing2024}. These help, but they are all attempts to make an unconstrained distribution stay near its training manifold by training harder.

Our claim is that the choice of \emph{what recurs} matters more than how hard it is regularized. If the quantity fed back is an interpretable, metric state in place of an appearance latent, the drift becomes something a rule can act on, and three distinct kinds of leverage appear. (i)~Whole failure modes stop being representable. The bridge draws a fixed skeleton topology, so a monster cannot become a different species, and cannot gain or lose a limb. (ii)~The state can be checked against an external structure between steps. The terrain is a fixed, known object, so an infeasible configuration can be projected back onto the feasible set every frame, and the re-sampled patch means the next step is conditioned on the corrected state, a per-step contraction that no purely generative rollout has access to. (iii)~Violations are measurable in absolute units. Penetration depth, foot-skate, and bone-length variation are quantities with metres attached, so drift can be detected and attributed.
That is what lets us report that a smooth FVD curve can coexist with a state-layer failure the
curve does not register, which is exactly what the drift of \S\ref{sec:exp:ec} shows.

This is a familiar bargain in character animation, where controllers have long been conditioned on the local terrain profile and constrained by a physics engine rather than asked to learn ground contact from data alone \citep{holden2017pfnn,peng2018deepmimic}. What the offloading view adds is that the same reasoning applies to a \emph{world} model, and that it gives a criterion for choosing the intermediate representation. A representation
justifies its cost when rules can be enforced on it. Compactness and reconstruction
fidelity, the usual criteria, say nothing about that. The negative result above marks where the argument stops. Rules help where the model cannot
trade against them.

\section{Dynamics Model Details}
\label{app:dynamics}
Both stages are causal transformers with sinusoidal position encodings and a bounded attention window (block size $512$). \textbf{ActionGPT} (decision): $4$ layers, width $256$, $8$ heads, $\sim$$2.5$M parameters. Per frame it consumes a low-dimensional state summary: the $18$D root and rotation stream (both entities' root displacement and 6D rotation), learned per-entity action embeddings, animation-progress scalars, and the weapon sub-state.
From these it predicts the next root displacement and rotation, together with a categorical distribution over each entity's next action id. Optional terrain conditioning (an egocentric height patch together with per-joint ground clearances, encoded by a $2$-layer MLP) is concatenated to the input. \textbf{PoseGPT} (animation): $8$ layers, width $512$, $8$ heads, $\sim$$25$M parameters. It consumes the recent $258$D body state and the (possibly overridden) action ids and predicts the next-frame $258$D body pose, and the action input is aligned to the frame being animated (a one-step shift relative to the decision stream). Terrain-aware training adds ground-contact and non-penetration objectives so that predicted joints respect the local surface. \textbf{Control injection.} At inference, replacing an entity's sampled action id with a target id at a given frame steers that entity. This is a single tensor assignment with no gradient or auxiliary network, and is the mechanism used by the controllability probe (Appendix~\ref{app:controllability}).

\textbf{Two control channels: token and root.} Control is not confined to the action vocabulary. ActionGPT regresses the root displacement and the 6D root rotation as continuous outputs alongside the action logits.
PoseGPT \emph{consumes} that root stream instead of deriving it from the body, so the root is an input to animation and can be overwritten at a frame exactly as an action token is. Writing a displacement fixes travel speed and direction, writing a rotation fixes heading, and the animation stage then articulates the body under the imposed root. The two channels control different quantities and can be used together. An action id says \emph{what} the character is doing and is the natural interface for discrete, button-like commands, but it \emph{steers} rather than pins, because the pose a token produces still depends on the context it is injected into (Appendix~\ref{app:controllability}). Overwriting the root pins locomotion exactly, at the price of saying nothing about the limbs, so a body whose held action has finished animating will translate without stepping unless the action channel is driven as well. Our movement demonstrations therefore script the root while leaving the action stream free, our action demonstrations do the converse, and both remain single tensor assignments on the model's own output stream. Because the root is metric and integrated by the bridge rather than by the network (Appendix~\ref{app:bridge}), a root command is expressed in metres and radians and is reproduced exactly in world space.

\section{Evaluation Protocol Details}
\label{app:protocol}
\begin{figure}[t]
\centering
\includegraphics[width=\linewidth]{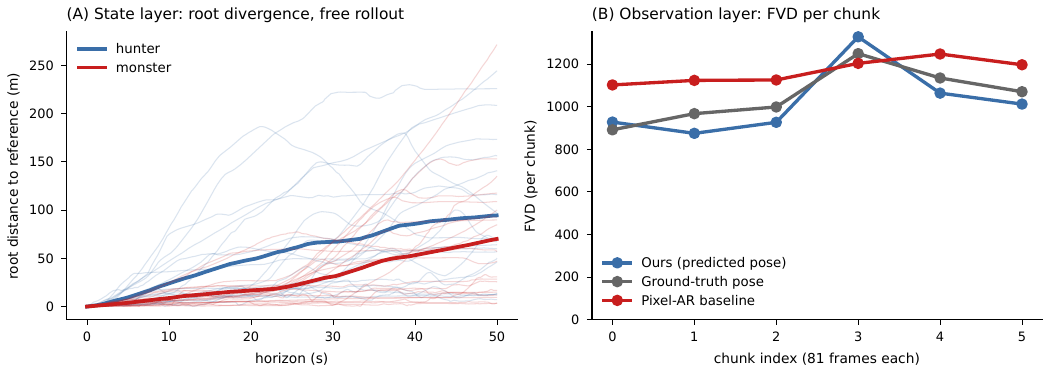}
\caption{Long-horizon behaviour on both evaluation layers (\S\ref{sec:exp:ec}). (A)~State layer:
root distance to the recorded reference for $20$ held-out free rollouts (faint) and their means (bold). A plausible stochastic world legitimately diverges from any single reference, so this panel is a diagnostic of divergence in absolute units and not a score. (B)~Observation layer: FVD per $81$-frame chunk rises slowly and smoothly for both models, and does not separate them.}
\label{fig:longhorizon}
\end{figure}
\paragraph{State-layer metrics.}
\emph{RA-MPJPE} adapts the standard mean per-joint position error \citep{h36m2014} by subtracting each entity's root before comparing joints, isolating articulated-pose error from global translation. \emph{Root drift} is the Euclidean distance between predicted and reference root position as a
function of horizon; the drift curves of Figure~\ref{fig:longhorizon} use twenty held-out free
rollouts of $1{,}000$ frames at $20$\,fps from a $64$-frame seed context, a longer state-only
horizon than the twelve rendered windows. \emph{Bone-length stability} is the coefficient of variation of each bone length over a rollout (a rig-consistency check). \emph{Foot-skate} measures horizontal motion of a foot joint while it is in ground contact, using a rolling-window local floor estimate (window $1$\,s, contact within $8$\,cm of the local
floor) and a vertical-velocity gate ($<3$\,cm per frame) to define contact; a global percentile
floor fails for roaming characters. The recorded-state value is reported beside every
generated one, so only the comparison is read. \emph{Ground penetration} measures joints below the local surface.
\paragraph{Observation-layer metrics.}
We report FVD \citep{fvd2018} for perceptual and temporal quality. For long rollouts we report FVD per chunk against the chunk index (\S\ref{sec:exp:ec}). Beyond the seed chunk we score distributionally rather than per frame, since a stochastic world may legitimately diverge from the one recorded reference and reference-based scores (PSNR, LPIPS \citep{lpips2018}) would charge it for doing so.
\paragraph{Controllability.}
We use the state-layer forced-action probe of Appendix~\ref{app:controllability}.

\section{State-Layer Controllability: Full Protocol and Results}
\label{app:controllability}
We test whether forcing an entity's action token moves the generated pose. For each held-out segment we seed the two-stage model with a common context and then roll out for a fixed horizon under three action streams for the hunter entity, with everything else identical. Under \textsc{Free} the model samples actions itself, under \textsc{Force-GT} the actions are overridden with the ground-truth ids, and under \textsc{Force-Shuf} they are overridden with a temporally shuffled and therefore mismatched GT stream. We report hunter RA-MPJPE against the ground-truth pose, and because the root is aligned away this measures whether the \emph{action shape} follows the forced token. We decode deterministically (temperature $0$) so the three conditions differ only in the injected token stream.

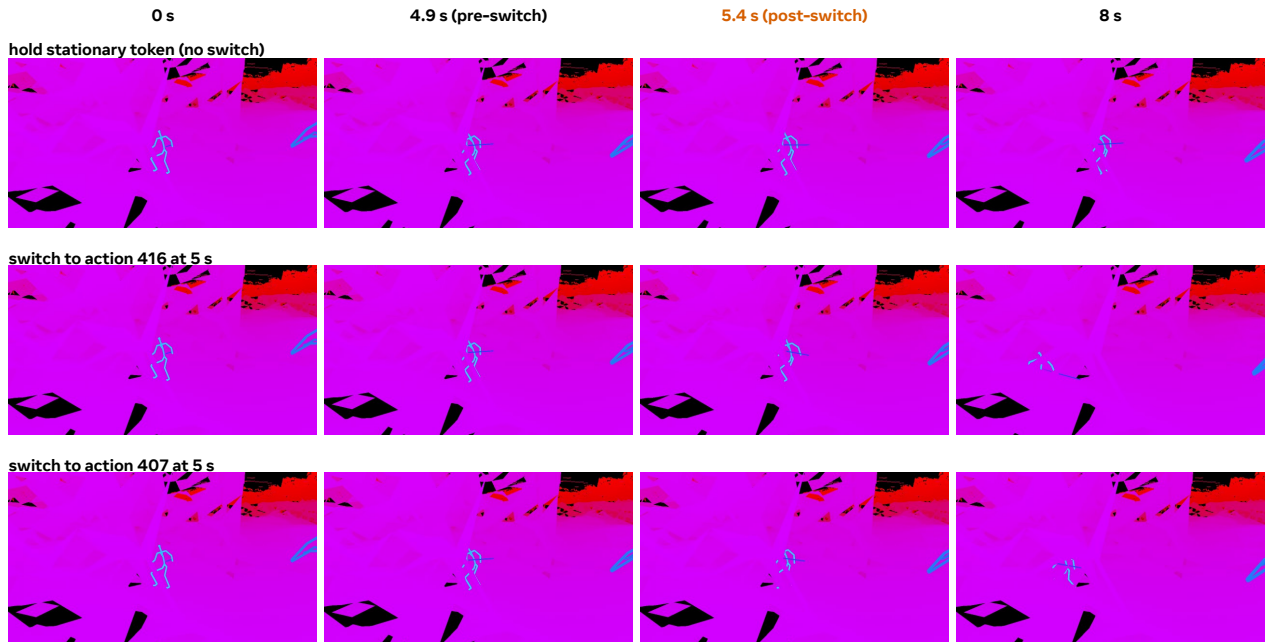
\begin{figure}[t]
\centering
\input{figures/panels/ctrl_timing.tex}
\caption{Token switching executes at the commanded frame (state layer,
pose-control rendering, fixed camera). All three rollouts share the seed and the sampling seed, so the frames are pixel-identical until the switch. At $5$\,s the hunter's token is held (top) or replaced by two different action ids (middle, bottom), and the generated pose diverges accordingly within a few frames.}
\label{fig:ctrltiming}
\end{figure}

\begin{figure}[t]
\centering
\includegraphics[width=0.6\linewidth]{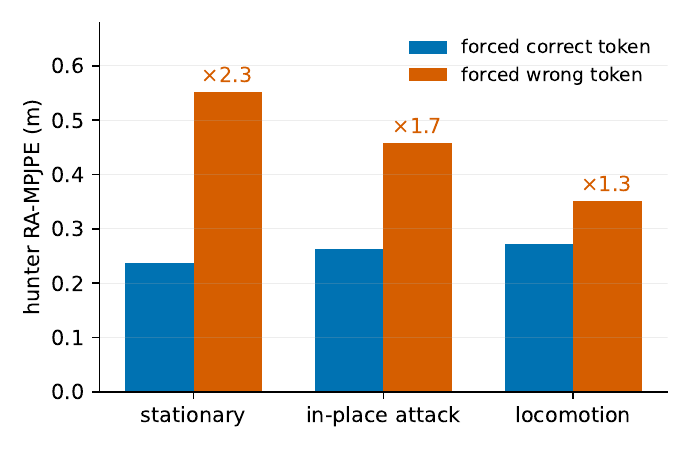}
\caption{Forced-token compliance by action class ($48$ held-out segments,
deterministic decoding). Error inflation from a wrong token is largest for the stationary and attack classes and
smallest for locomotion, whose pose varies least across ids. Locomotion ids differ least in pose shape and show the smallest inflation.}
\label{fig:perclass}
\end{figure}

Table~\ref{tab:controllability} reports the probe over $48$ held-out segments. The probe shows two distinct effects. First, forcing a \emph{mismatched} stream degrades the pose by $31\%$ ($0.272 \to 0.357$ mean RA-MPJPE, with the pairwise ordering $\textsc{Force-GT}<\textsc{Force-Shuf}$ holding on $33/48$ segments), which shows the token has causal authority, since the generated body executes the wrong action. Second, $\textsc{Force-GT}$ is only marginally better than $\textsc{Free}$ ($0.272$ against $0.281$), which says that forcing the correct token helps only where the free rollout would have chosen
differently, and on held-out segments it seldom does. For an interactive system this is the desired asymmetry, since a player's input matters precisely when it differs from what the world would have done anyway. The per-class breakdown shows compliance is sharpest for the action classes a player actually toggles, since forcing a wrong token inflates error by $2.3\times$ for stationary actions and $1.8\times$ for in-place attacks,
against $1.3\times$ for locomotion, whose pose shape varies least across ids.
\begin{table}[h]
\centering
\caption{State-layer controllability probe at scale. Hunter root-aligned MPJPE (m)
at horizon $120$, seed context $64$ frames, deterministic decoding; $48$ held-out segments (mean $\pm$ std). Bottom: per-frame error bucketed by the class of the driving token.}
\label{tab:controllability}
\begin{tabular}{lccc}
\toprule
 & \textsc{Free} & \textsc{Force-GT} & \textsc{Force-Shuf} \\
\midrule
RA-MPJPE (m) & $0.281 \pm 0.141$ & $\mathbf{0.272 \pm 0.135}$ & $0.357 \pm 0.120$ \\
\midrule
per class (\textsc{GT}$\to$\textsc{Shuf}) & \multicolumn{3}{c}{%
stationary $0.24\!\to\!0.55$ \quad attack $0.26\!\to\!0.46$ \quad locomotion $0.27\!\to\!0.35$} \\
\bottomrule
\end{tabular}
\end{table}
Figure~\ref{fig:ctrltiming} shows the probe's mechanism directly on the state layer: three rollouts share an identical seed and an identical prefix, differ only in the token injected from the $5$-second mark, and the generated body executes the respective action within a few frames of the switch. Figure~\ref{fig:perclass} plots the per-class breakdown of Table~\ref{tab:controllability}.

\section{Baseline Scope}
\label{app:baseline}
We restrict the head-to-head to an end-to-end pixel-autoregressive model trained on our data and conditioning because a comparison is only interpretable when the task is held fixed. Many interactive video world models evaluate under protocols that differ from ours in ways that preclude a direct numerical comparison.
Examples are a fixed single camera, a single controllable entity with the rest left as passive scenery, and control inputs such as device signals or per-engine animation ids that our data does not provide. Forcing such systems onto our protocol would require re-implementing and re-training them under assumptions they were not designed for, and any resulting number would reflect that mismatch. We therefore compare against the baseline that isolates our actual claim, namely \emph{where} the world state is represented, and we position the broader landscape qualitatively in \S\ref{sec:related}.

\begin{figure}[h]
\centering
\input{figures/panels/closed_panel.tex}
\caption{Qualitative comparison under one timestamped action schedule (stand
still until $5$\,s, then one large melee attack), all systems starting from the same recorded first frame. Our model receives the schedule as injected action tokens. The baseline and the commercial systems receive it as text. Commercial systems render superb appearance but follow the schedule loosely or not at all (Seedance never attacks; Grok attacks seconds late and swaps both characters' appearance), and the pixel-autoregressive baseline exchanges the monster's identity. Precise entity-level control at a commanded time is the axis the decoupled state interface provides.}
\label{fig:closedpanel}
\end{figure}
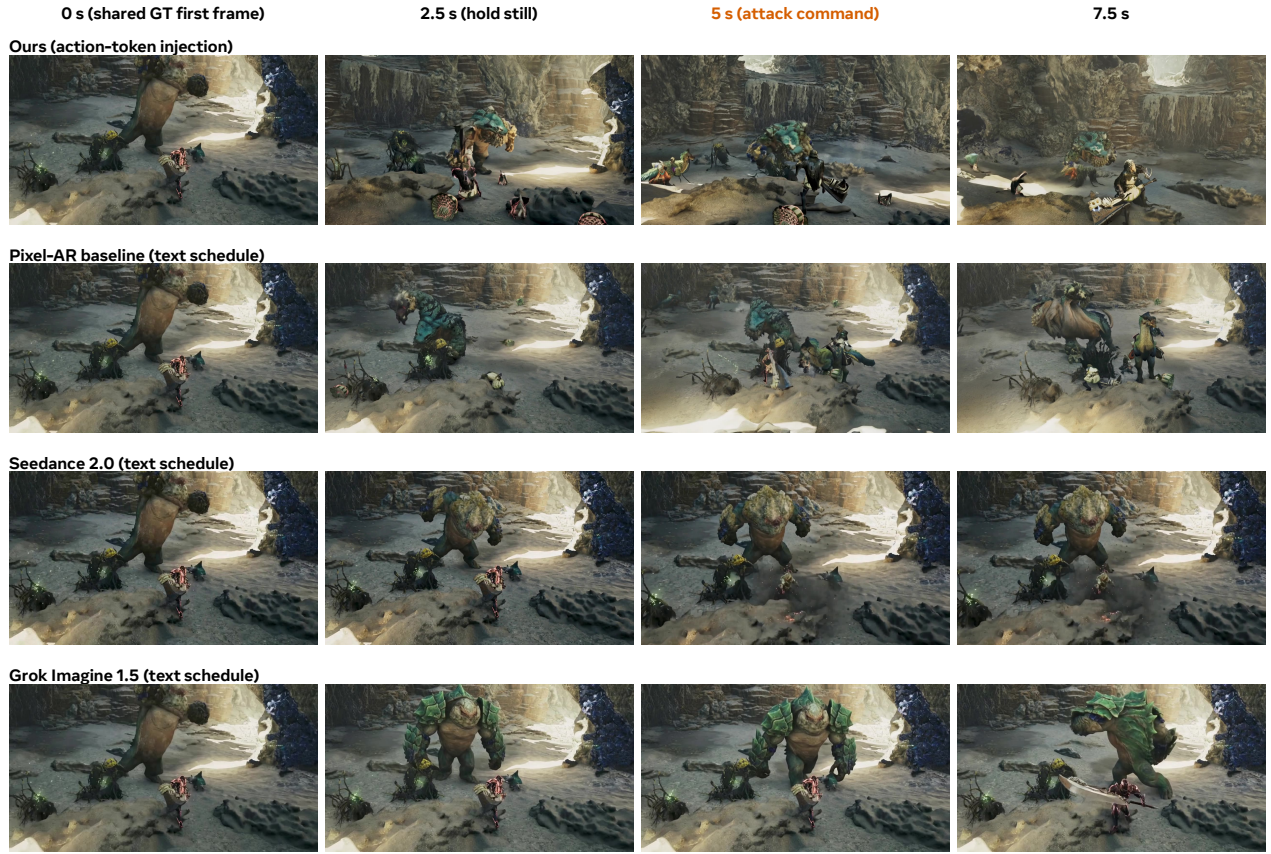

\paragraph{Commercial general-purpose video generators.}
A second class of systems enters as a qualitative reference: the commercial general-purpose video generators. These models produce high-fidelity clips from a text prompt and an optional first frame, but they expose no interface for the structured, per-frame control our setting is about, since they accept neither an articulated pose stream, nor an action-token schedule, nor a first-frame plus control-reference pairing. A metric that scores action following or long-horizon state consistency therefore has no matched input to give them, and any number we reported would measure the protocol mismatch. Following common practice for game world models in comparably complex settings, we compare
against these systems \emph{qualitatively}.
Each is driven with the same first frame and a normalized action prompt, in which the intended events are written as an explicit timestamped schedule so that a model's built-in prompt rewriting cannot silently change the task.
The clips are presented side by side.
Open-ended visual quality is no longer the bottleneck for these systems, while precise control over specific entities still is, and that is the capability an explicit state provides. Quantitative tables (\S\ref{sec:exp}) are reserved for models we run under our own protocol,
which at present is the pixel-autoregressive baseline above. Figure~\ref{fig:closedpanel} shows the comparison for one action schedule, with Seedance~2.0 \citep{seedance2_2026} and Grok~Imagine~1.5 \citep{grokimagine2026} as the commercial systems. Each receives the schedule described above (the full template is released with our code). The pattern matches the motivation above. Both systems produce high-fidelity footage and respect the first frame.
Seedance keeps the scene and characters faithful; Grok executes the attack roughly three
seconds late and meanwhile exchanges both characters' appearance. Neither offers a handle on \emph{which} entity does \emph{what} at \emph{which} time, whereas the decoupled model executes the token switch at the commanded frame. Seedance returns only $5$\,s of video, so its last column repeats its final frame.

\section{Licenses and Use of Assets}
\label{app:licenses}
We list every external asset this work uses, together with its verified license terms.

\paragraph{Wan2.2-Fun-5B-Control.}
We use the Wan2.2-Fun-5B-Control checkpoint as the backbone of our observation model. The weights are released by Alibaba-PAI under the Apache License 2.0, at \url{https://huggingface.co/alibaba-pai/Wan2.2-Fun-5B-Control}. The umT5-xxl text encoder and the CLIP (XLM-RoBERTa-large ViT-H) image encoder that our pipeline loads are distributed inside that same repository and under the same license, as is the Wan2.2 VAE.

\paragraph{VideoX-Fun.}
Our observation-model training and inference code extends the VideoX-Fun framework, released by aigc-apps under the Apache License 2.0, at \url{https://github.com/aigc-apps/VideoX-Fun}. Our dynamics model, deterministic bridge, terrain pipeline, and evaluation code will be released under the same license.

\paragraph{WildWorld.}
The gameplay recordings, the frame-level action ids, and the per-entity state we build on come from \textsc{WildWorld} \citep{wildworld2026}, a dataset we released previously and use here under the terms published with it. The scanned terrain height fields (Appendix~\ref{app:bridge}) are not part of that release. They are produced by this work and will be distributed with our code.

\paragraph{\emph{Monster Hunter Wilds}.}
\emph{Monster Hunter Wilds} (\copyright{} Capcom Co., Ltd.) is a commercial video game, and unlike the assets above it is not distributed under an open licence. All in-game footage used in this work was recorded by us, on retail copies of the game that we purchased and on our own player accounts, and is used solely for non-commercial academic research. Capcom publishes a \emph{Capcom Video Policy} (last updated March 15, 2021, \url{https://www.capcomusa.com/video-policy/}) which permits players to capture and share footage of its titles provided the use is not monetized outside the channels it enumerates. The policy also states explicitly that it is not itself an express permission or licence, and we do not claim one. The corpus this work builds on \citep{wildworld2026} contains recorded gameplay video, together with the derived artifacts we add here: state sequences, terrain height fields, metric implementations, and pose-control renders produced by our own bridge. All of it is made available for non-commercial academic research only, under the terms published with the release, and we ask that anyone using it observe the same restriction. We will promptly comply with any request from the rights holder concerning material released here.

\bibliographystyle{abbrv}
\bibliography{references}

\end{document}

%% file: figures/panels/bridge_strip.tex
\setlength{\tabcolsep}{1.5pt}
\renewcommand{\arraystretch}{0}
\begin{tabular}{@{}cccc@{}}
{\scriptsize\textbf{0 s (GT seed-end frame)}} & {\scriptsize\textbf{4 s}} & {\scriptsize\textbf{8 s}} & {\scriptsize\textbf{12 s}} \\[1pt]
\multicolumn{4}{@{}l@{}}{\rule{0pt}{2.4ex}{\scriptsize\textbf{pose-control video (deterministic bridge)}}} \\[0.5pt]
\includegraphics[width=0.2455\linewidth]{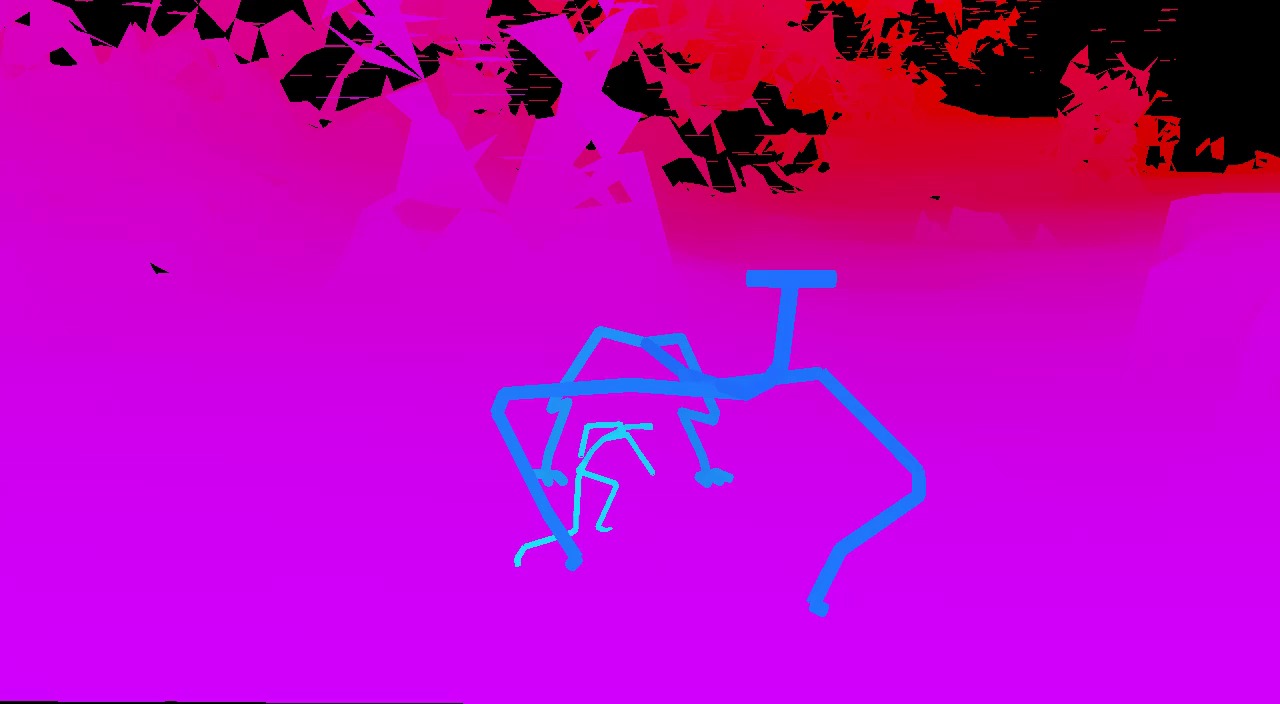} & \includegraphics[width=0.2455\linewidth]{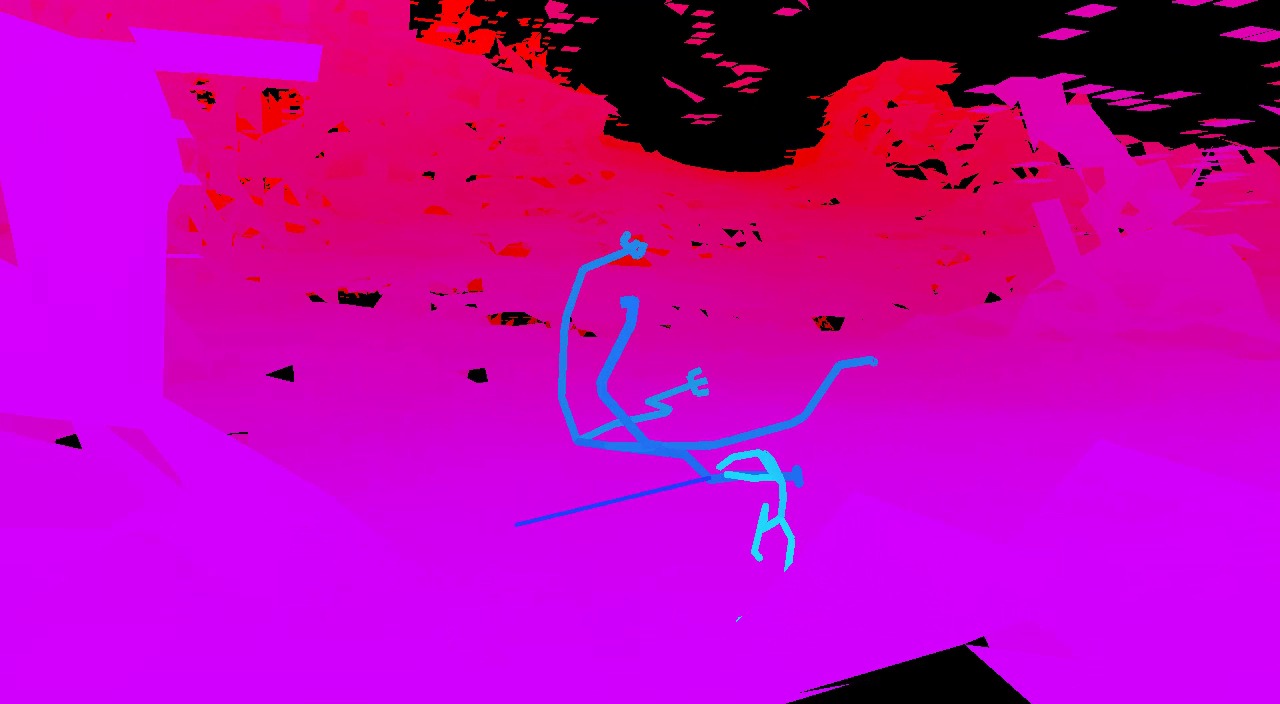} & \includegraphics[width=0.2455\linewidth]{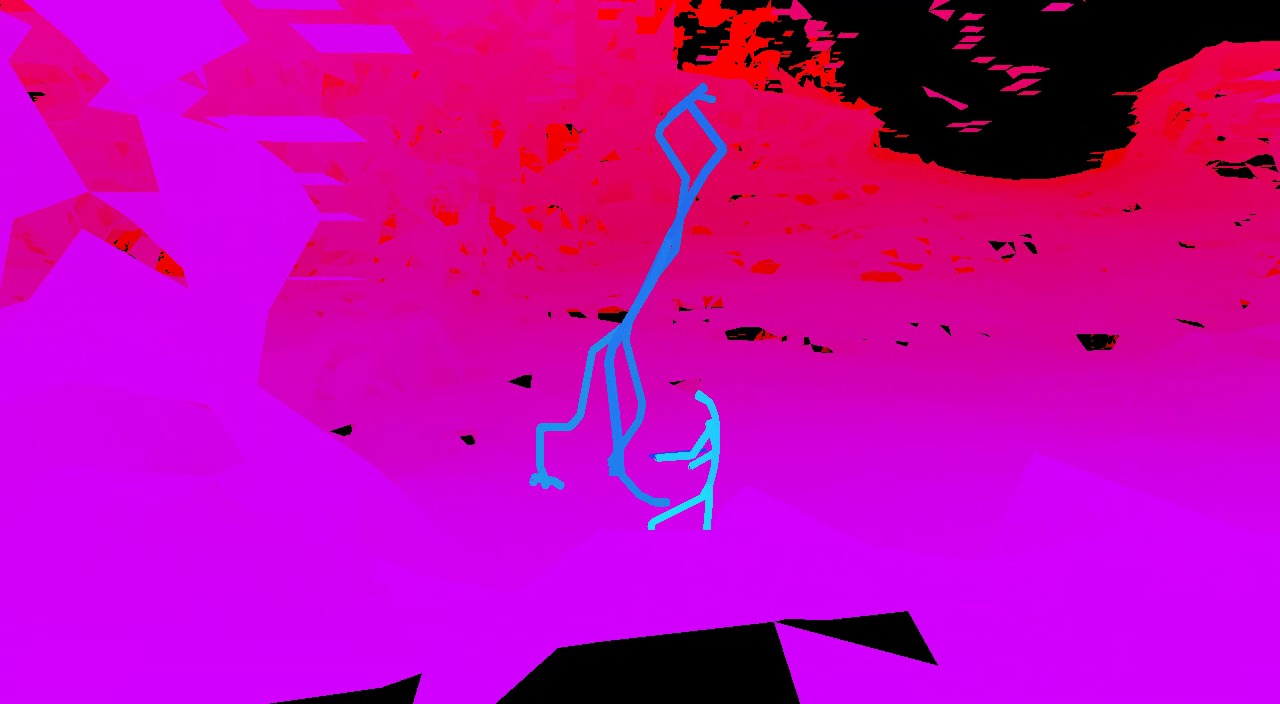} & \includegraphics[width=0.2455\linewidth]{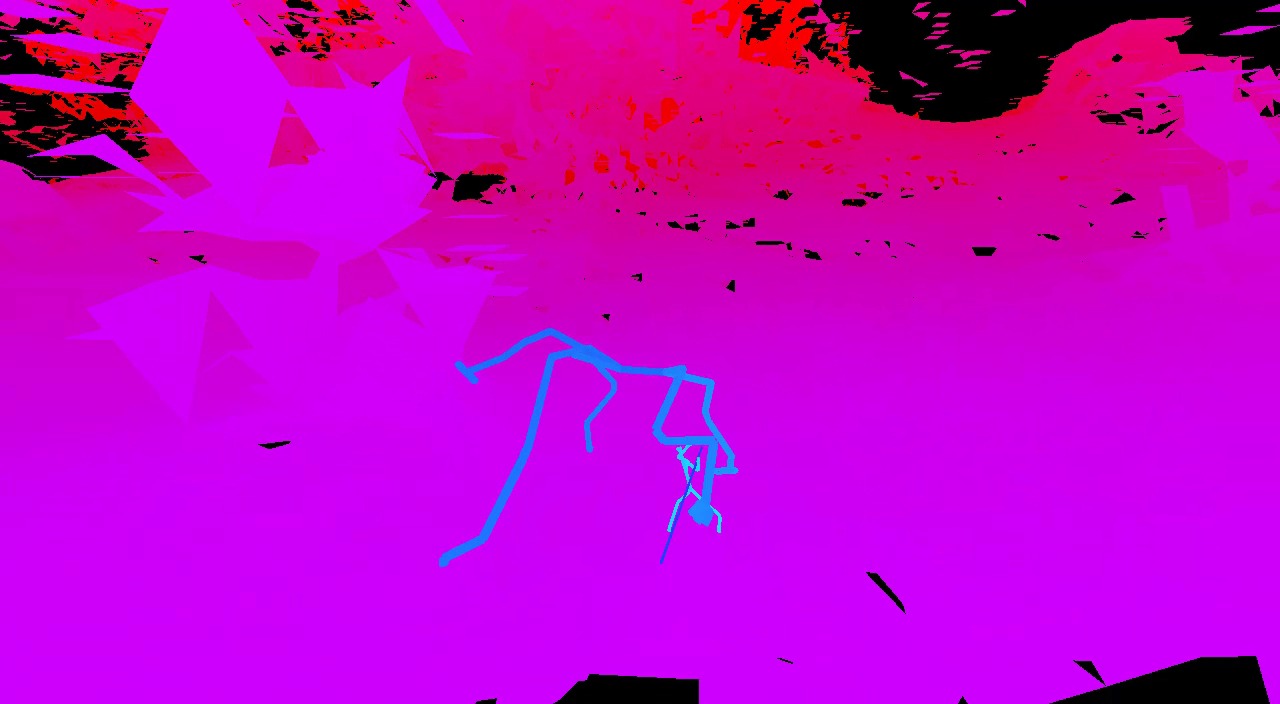} \\[2.5pt]
\multicolumn{4}{@{}l@{}}{\rule{0pt}{2.4ex}{\scriptsize\textbf{rendered RGB (observation model)}}} \\[0.5pt]
\includegraphics[width=0.2455\linewidth]{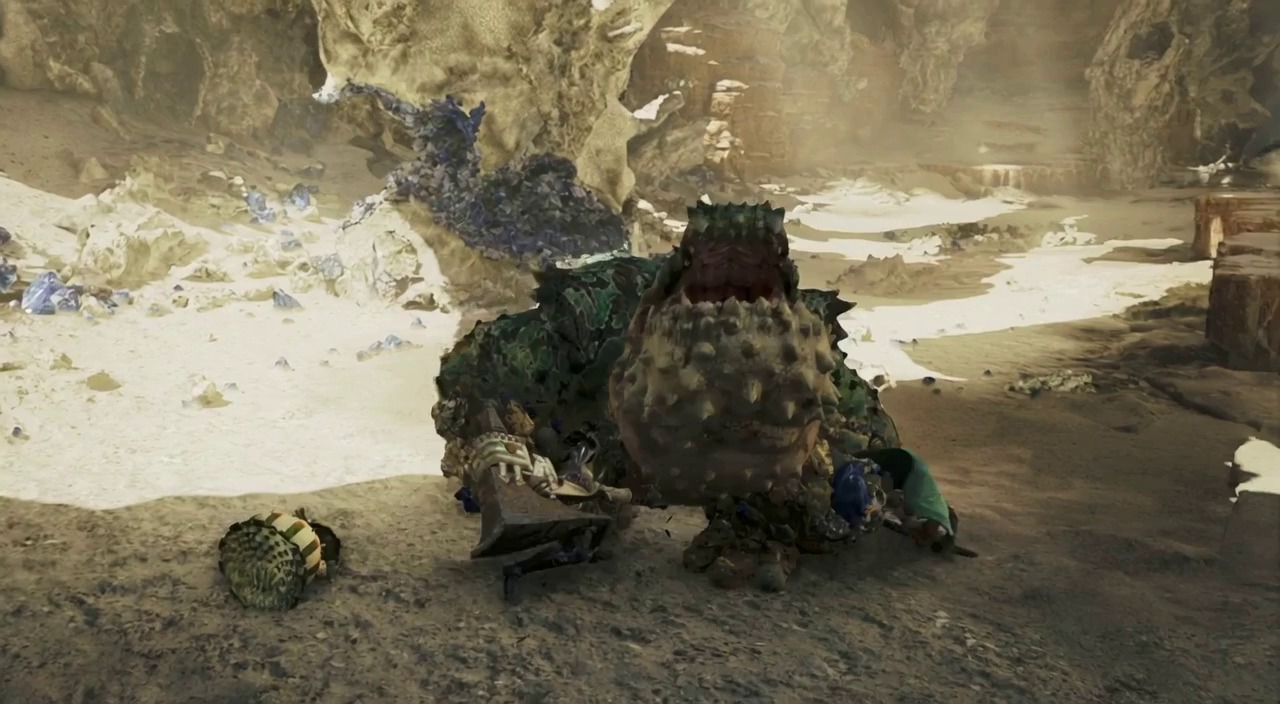} & \includegraphics[width=0.2455\linewidth]{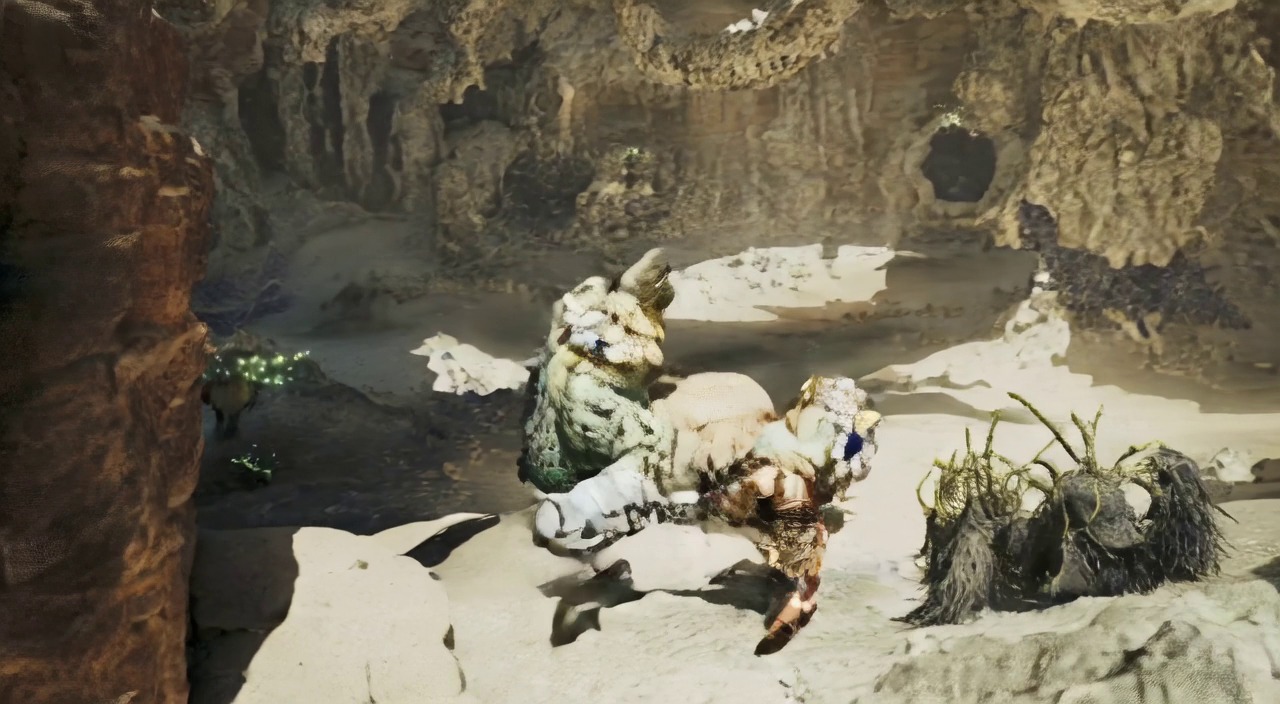} & \includegraphics[width=0.2455\linewidth]{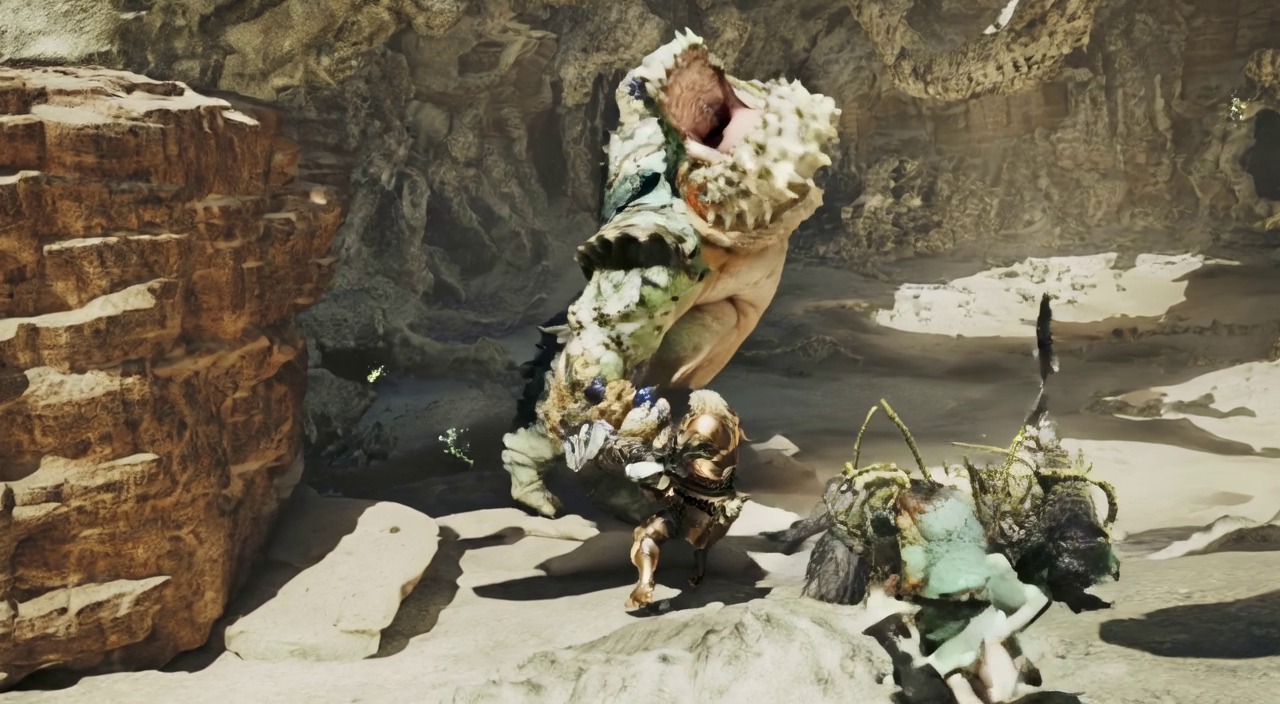} & \includegraphics[width=0.2455\linewidth]{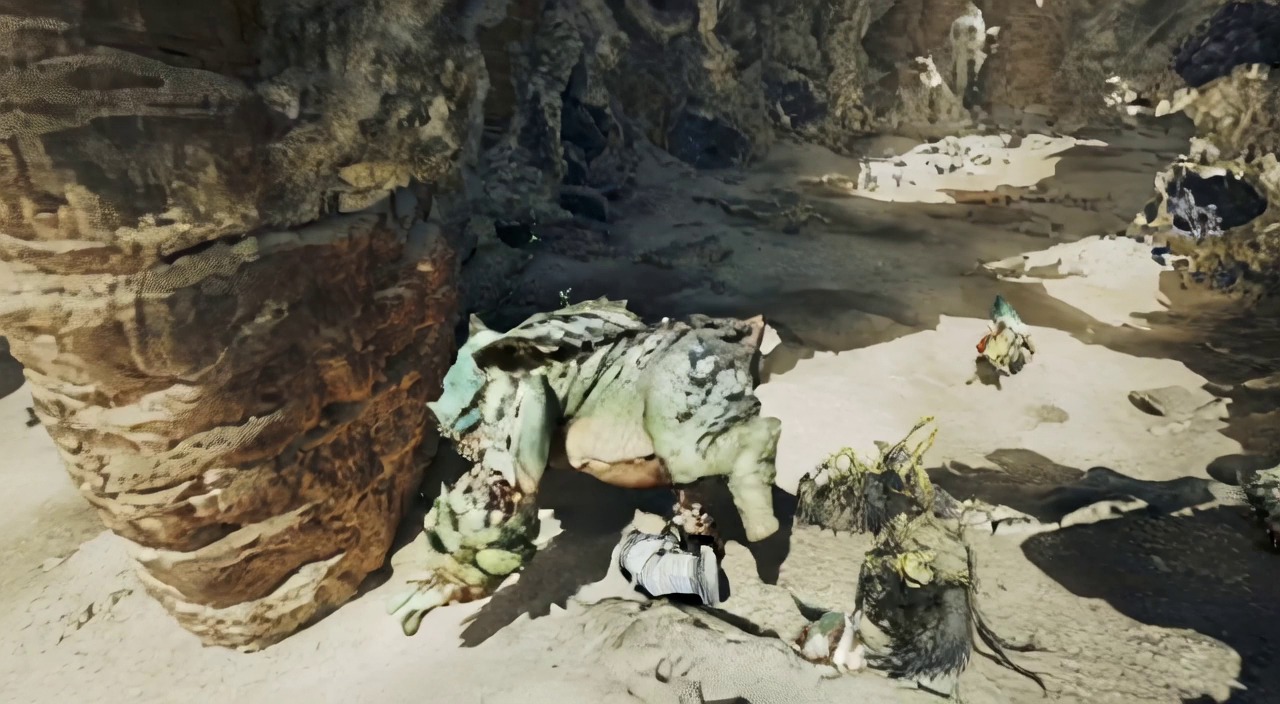} \\
\end{tabular}

%% file: figures/panels/ctrl_timing.tex
\setlength{\tabcolsep}{1.5pt}
\renewcommand{\arraystretch}{0}
\begin{tabular}{@{}cccc@{}}
{\scriptsize\textbf{0 s}} & {\scriptsize\textbf{4.9 s (pre-switch)}} & \textcolor{PanelAccent}{\scriptsize\textbf{5.4 s (post-switch)}} & {\scriptsize\textbf{8 s}} \\[1pt]
\multicolumn{4}{@{}l@{}}{\rule{0pt}{2.4ex}{\scriptsize\textbf{hold stationary token (no switch)}}} \\[0.5pt]
\includegraphics[width=0.2455\linewidth]{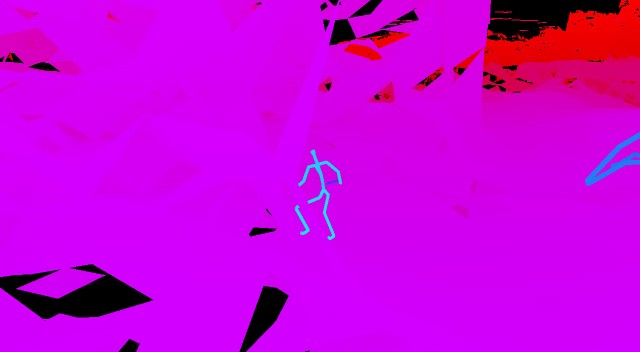} & \includegraphics[width=0.2455\linewidth]{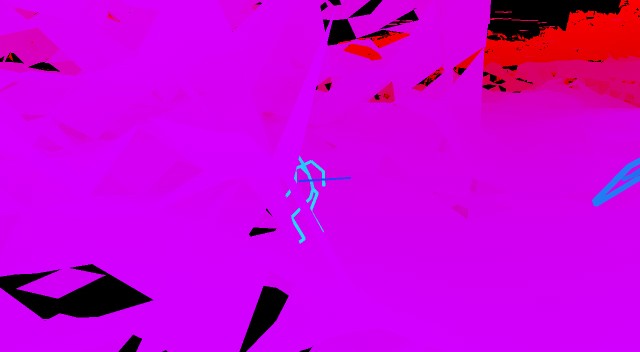} & \includegraphics[width=0.2455\linewidth]{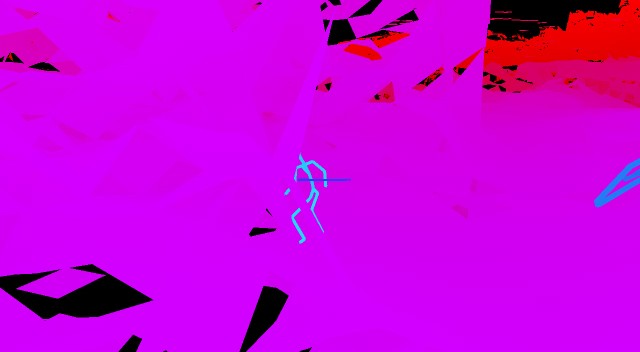} & \includegraphics[width=0.2455\linewidth]{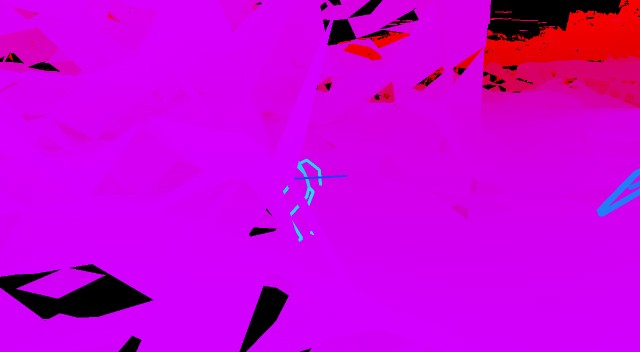} \\[2.5pt]
\multicolumn{4}{@{}l@{}}{\rule{0pt}{2.4ex}{\scriptsize\textbf{switch to action 416 at 5 s}}} \\[0.5pt]
\includegraphics[width=0.2455\linewidth]{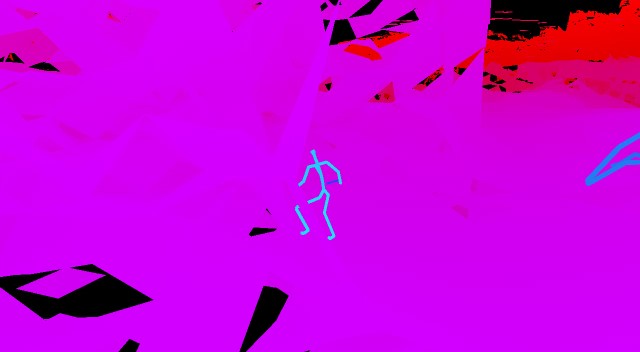} & \includegraphics[width=0.2455\linewidth]{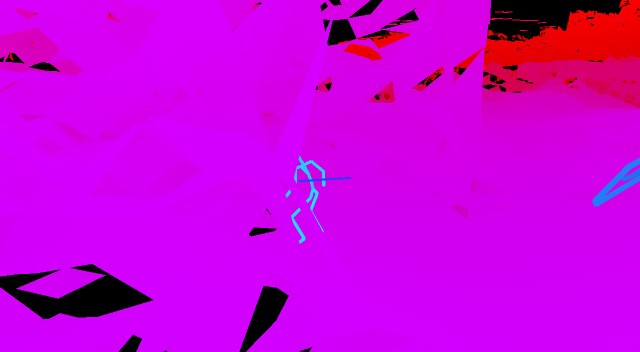} & \includegraphics[width=0.2455\linewidth]{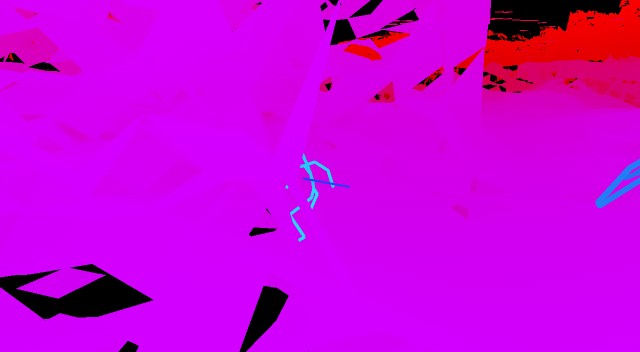} & \includegraphics[width=0.2455\linewidth]{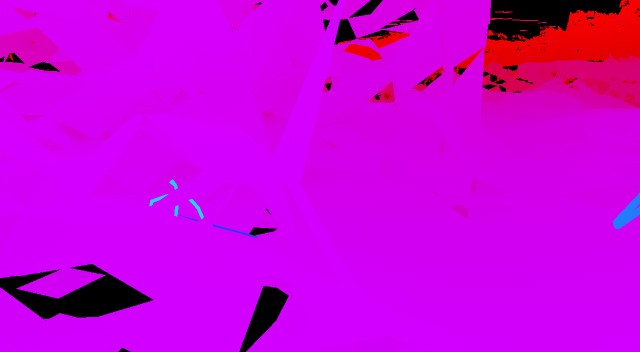} \\[2.5pt]
\multicolumn{4}{@{}l@{}}{\rule{0pt}{2.4ex}{\scriptsize\textbf{switch to action 407 at 5 s}}} \\[0.5pt]
\includegraphics[width=0.2455\linewidth]{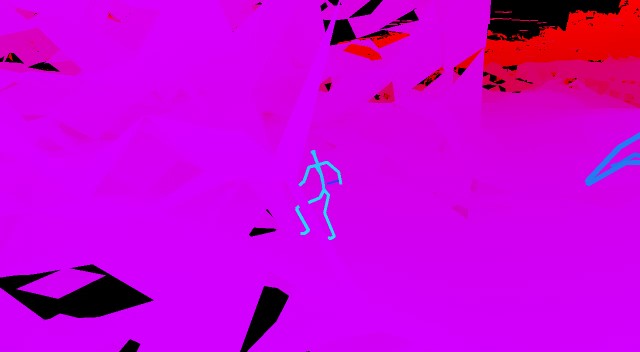} & \includegraphics[width=0.2455\linewidth]{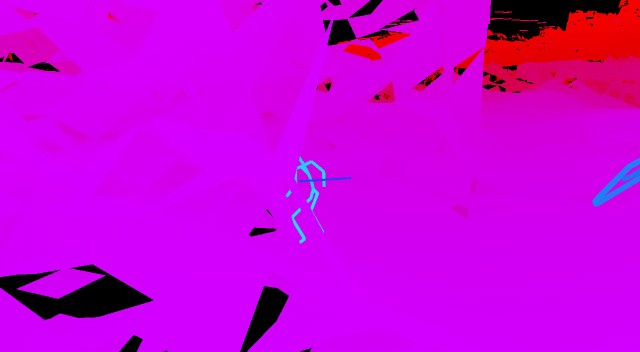} & \includegraphics[width=0.2455\linewidth]{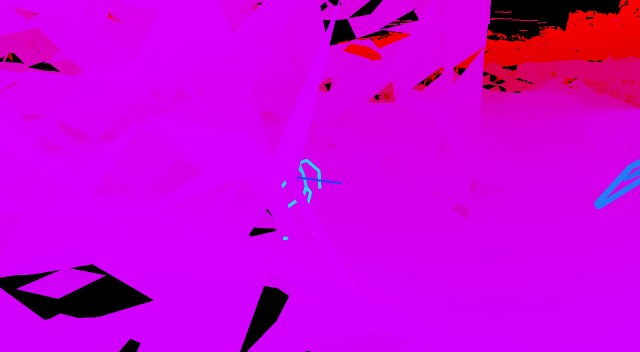} & \includegraphics[width=0.2455\linewidth]{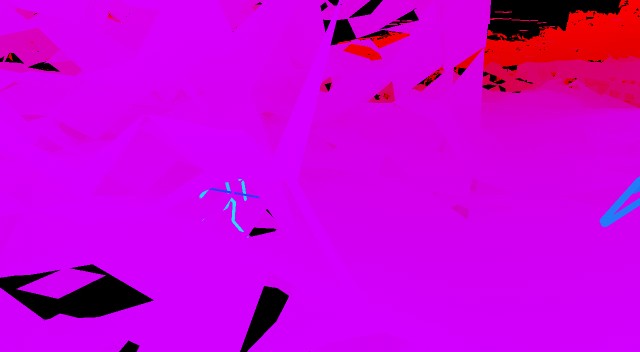} \\
\end{tabular}

%% file: figures/panels/closed_panel.tex
\setlength{\tabcolsep}{1.5pt}
\renewcommand{\arraystretch}{0}
\begin{tabular}{@{}cccc@{}}
{\scriptsize\textbf{0 s (shared GT first frame)}} & {\scriptsize\textbf{2.5 s (hold still)}} & \textcolor{PanelAccent}{\scriptsize\textbf{5 s (attack command)}} & {\scriptsize\textbf{7.5 s}} \\[1pt]
\multicolumn{4}{@{}l@{}}{\rule{0pt}{2.4ex}{\scriptsize\textbf{Ours (action-token injection)}}} \\[0.5pt]
\includegraphics[width=0.2455\linewidth]{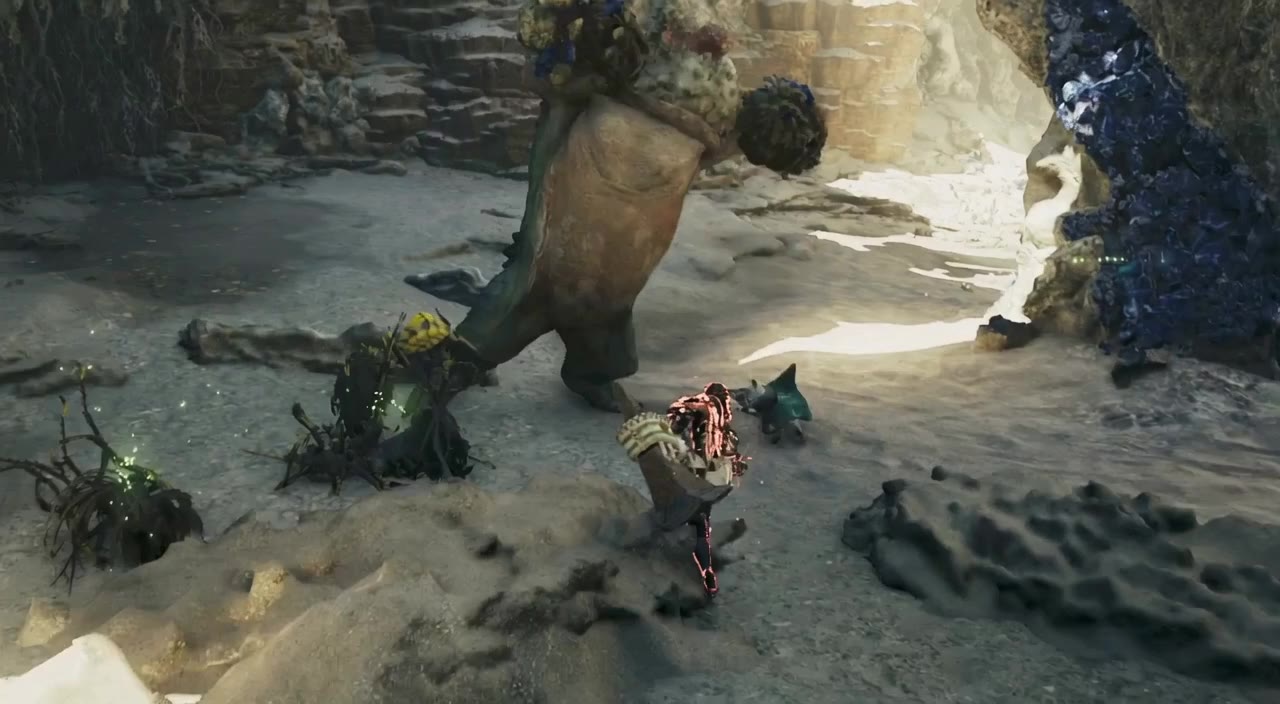} & \includegraphics[width=0.2455\linewidth]{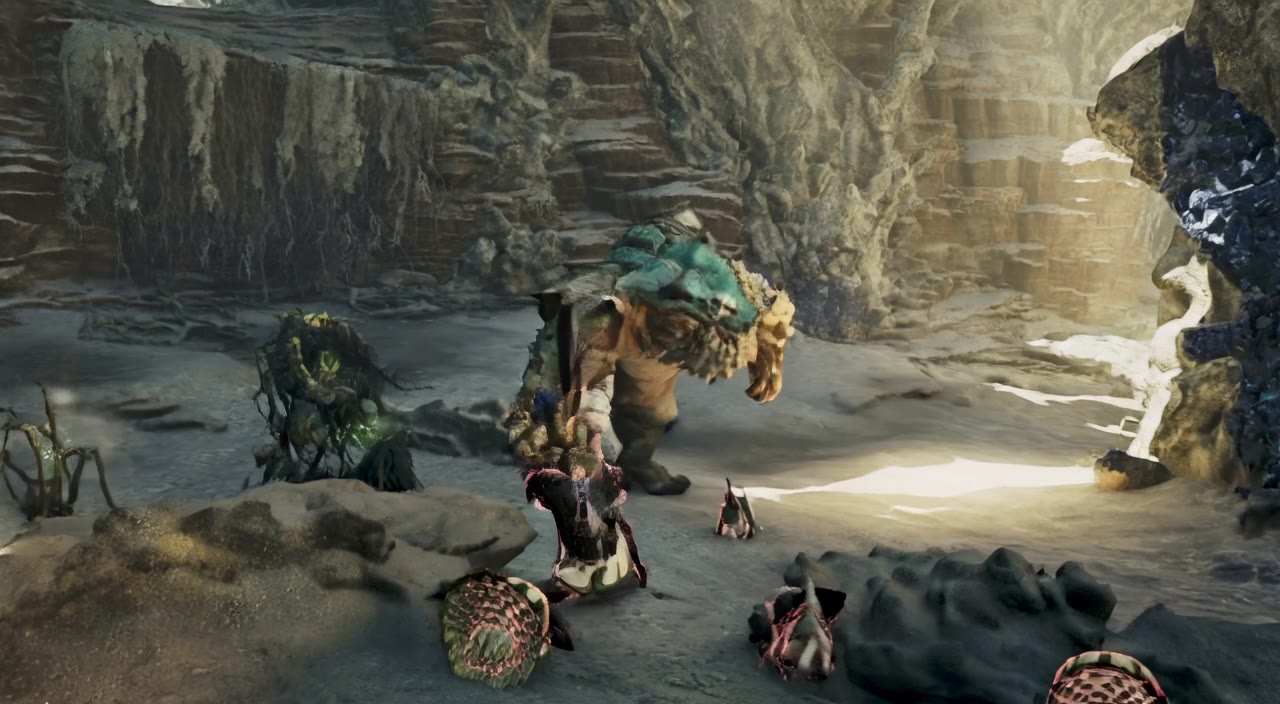} & \includegraphics[width=0.2455\linewidth]{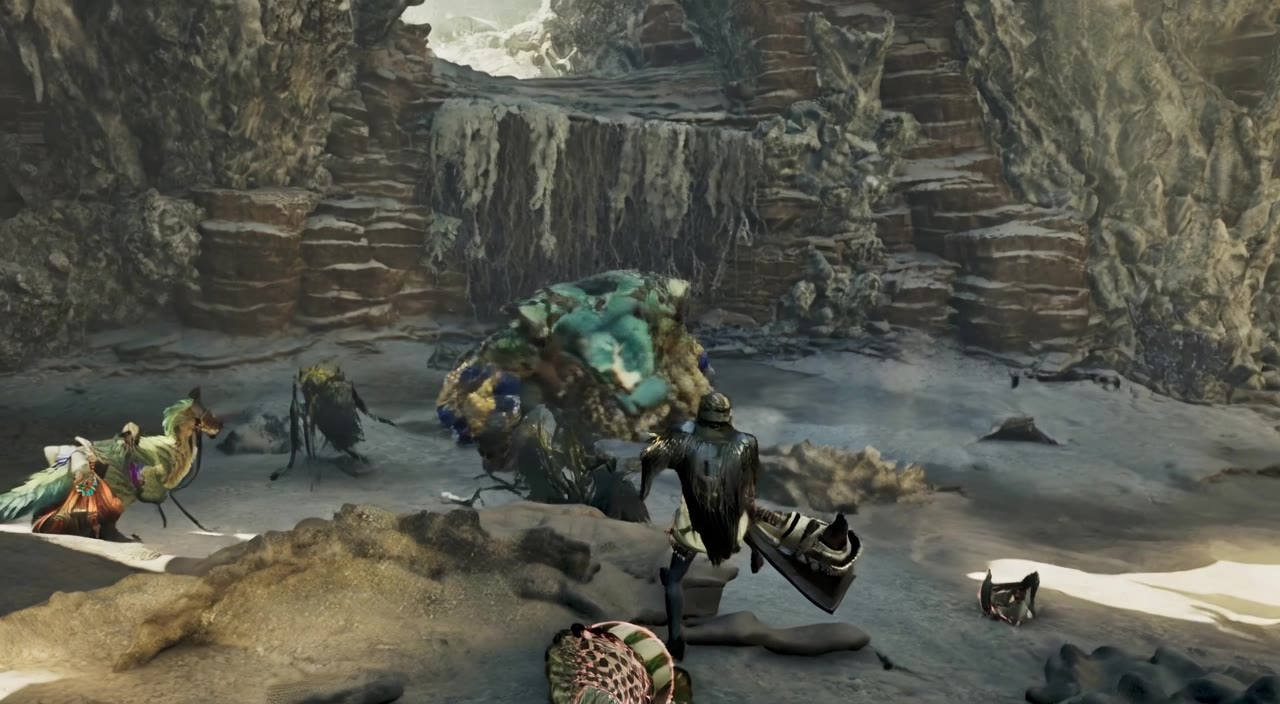} & \includegraphics[width=0.2455\linewidth]{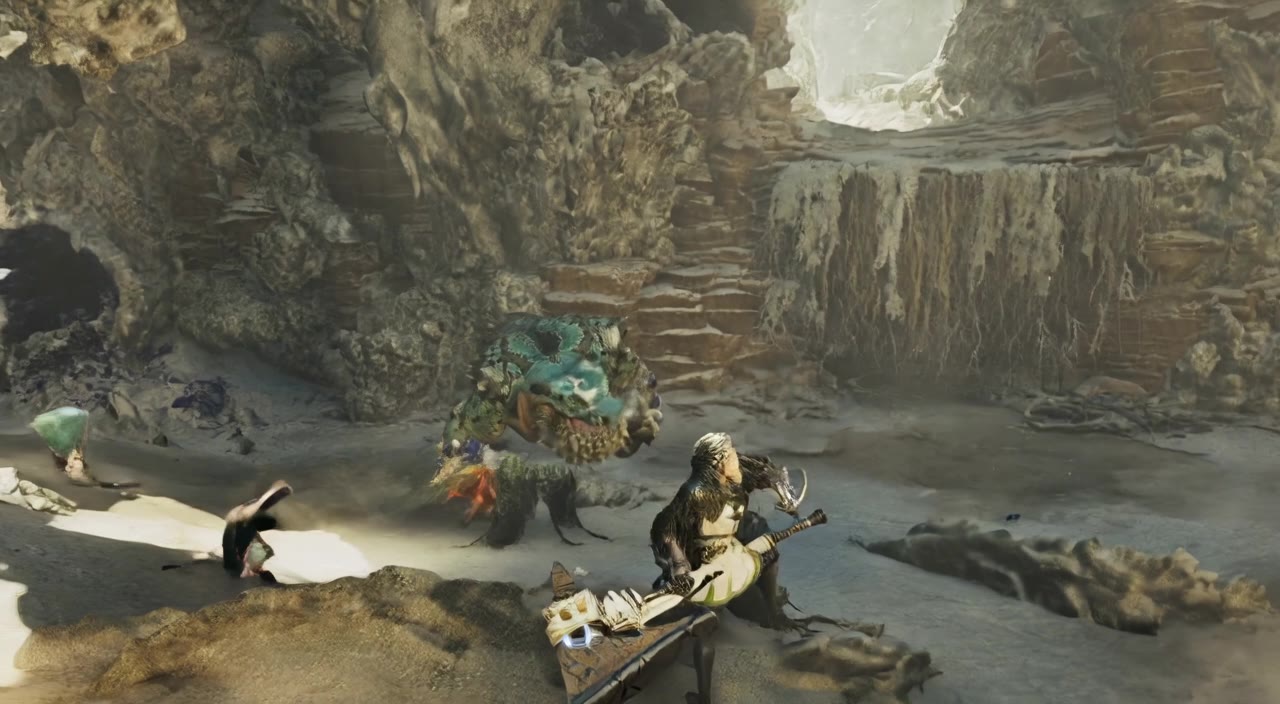} \\[2.5pt]
\multicolumn{4}{@{}l@{}}{\rule{0pt}{2.4ex}{\scriptsize\textbf{Pixel-AR baseline (text schedule)}}} \\[0.5pt]
\includegraphics[width=0.2455\linewidth]{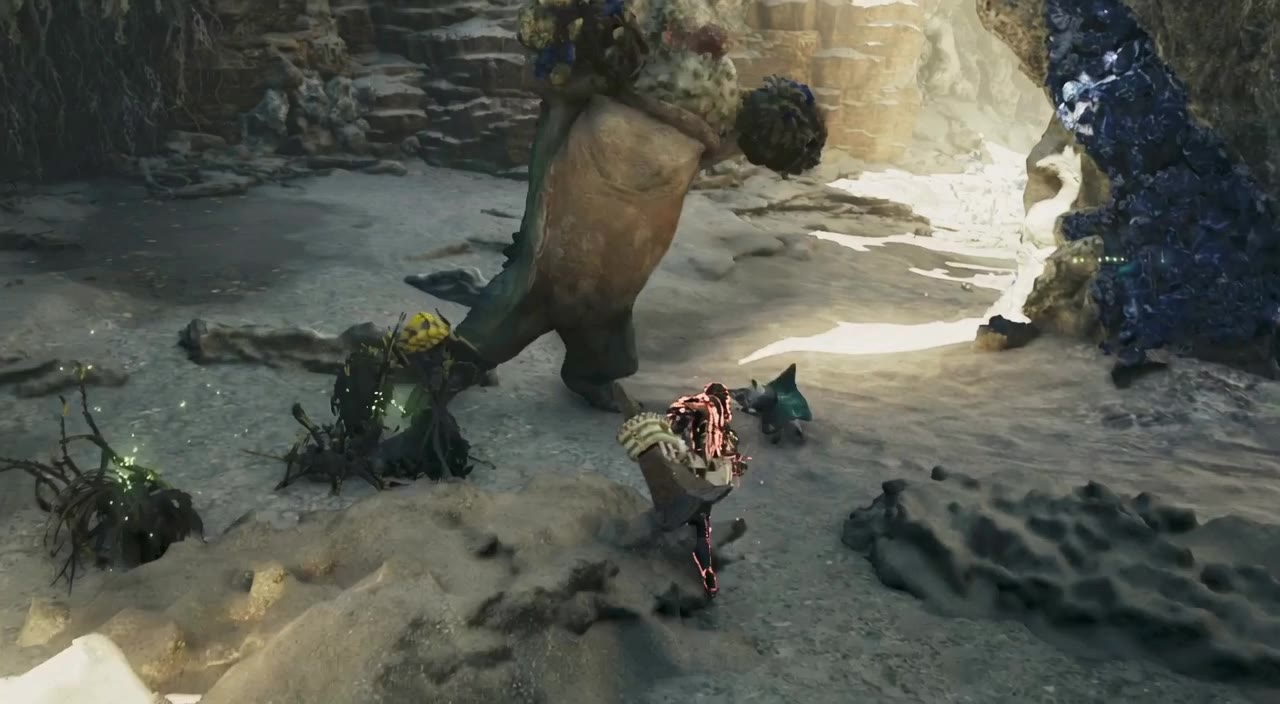} & \includegraphics[width=0.2455\linewidth]{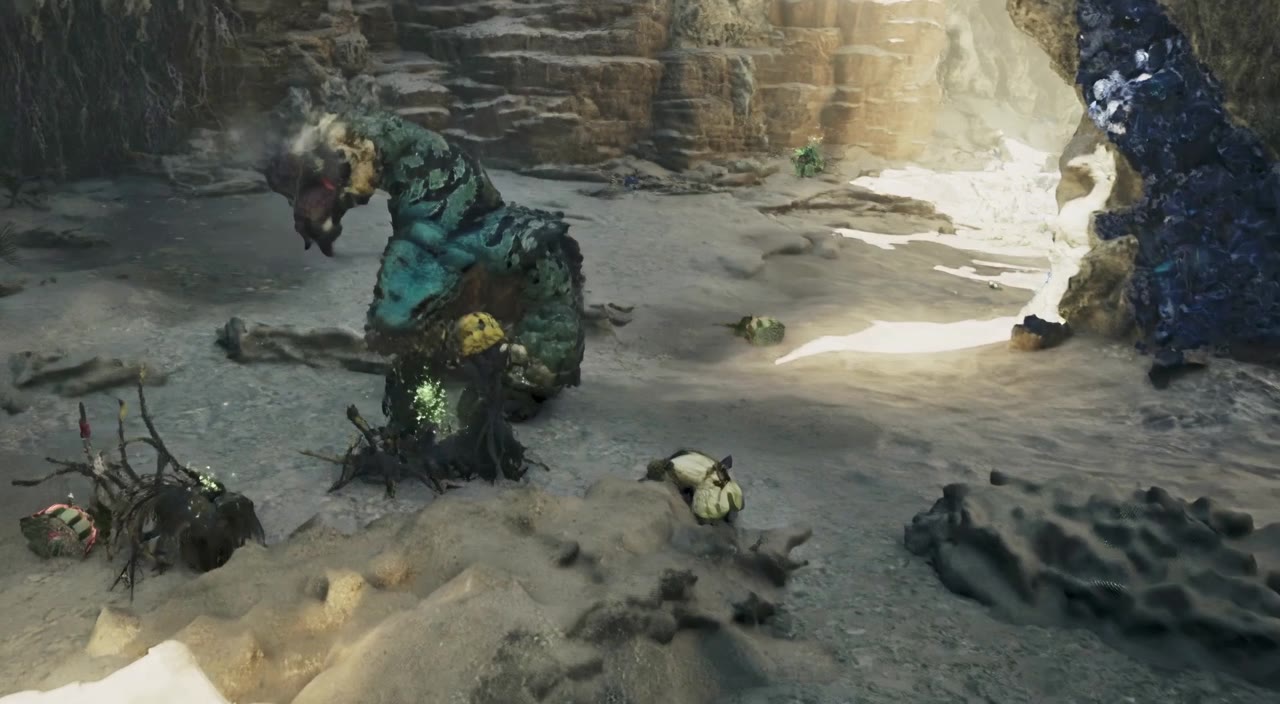} & \includegraphics[width=0.2455\linewidth]{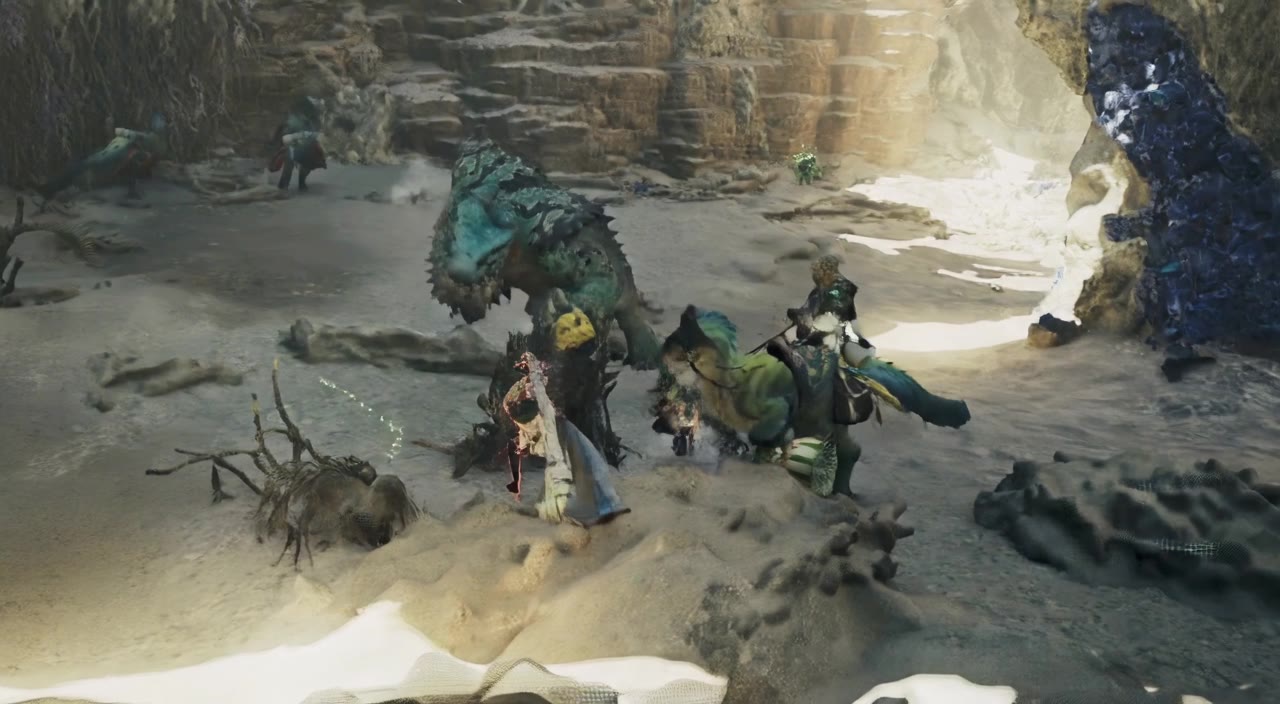} & \includegraphics[width=0.2455\linewidth]{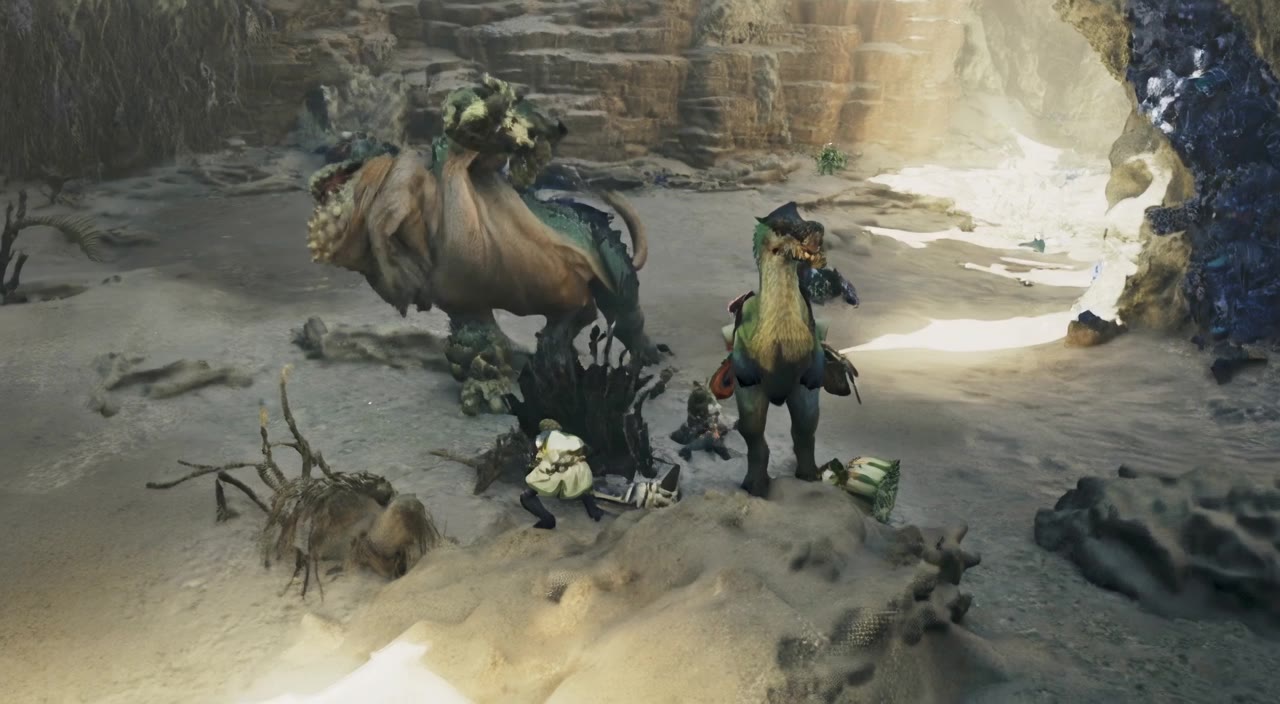} \\[2.5pt]
\multicolumn{4}{@{}l@{}}{\rule{0pt}{2.4ex}{\scriptsize\textbf{Seedance 2.0 (text schedule)}}} \\[0.5pt]
\includegraphics[width=0.2455\linewidth]{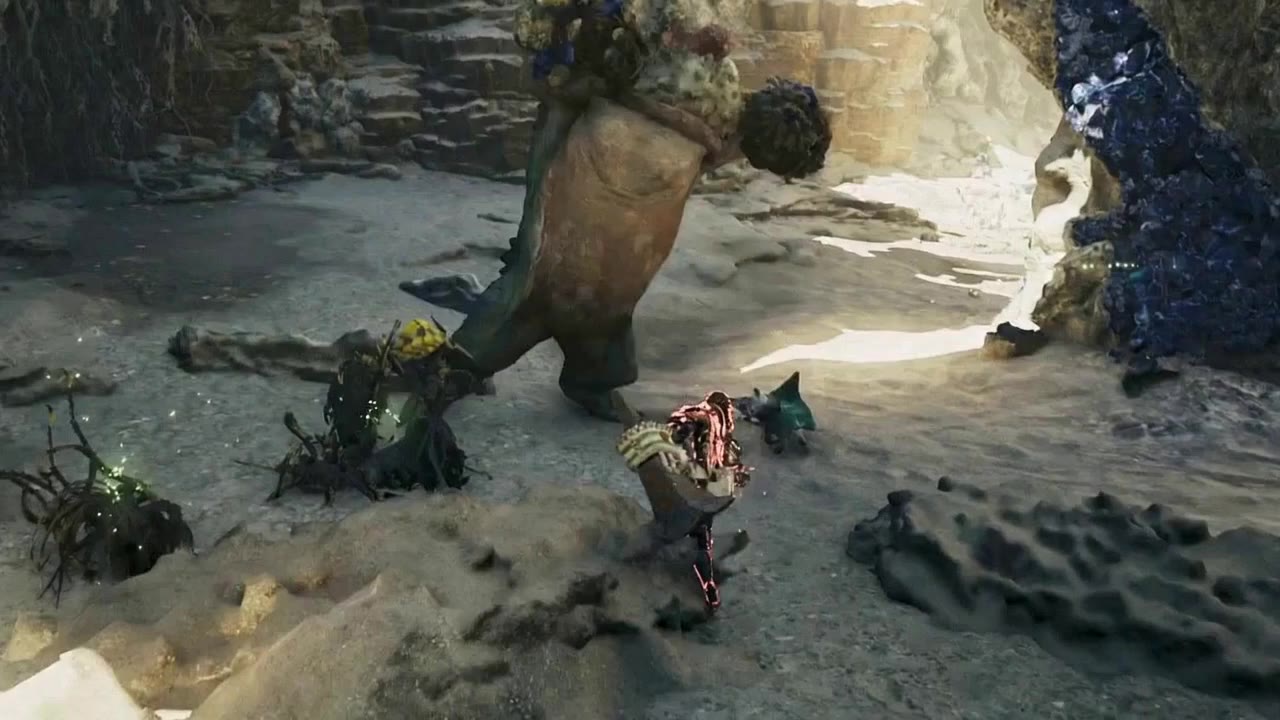} & \includegraphics[width=0.2455\linewidth]{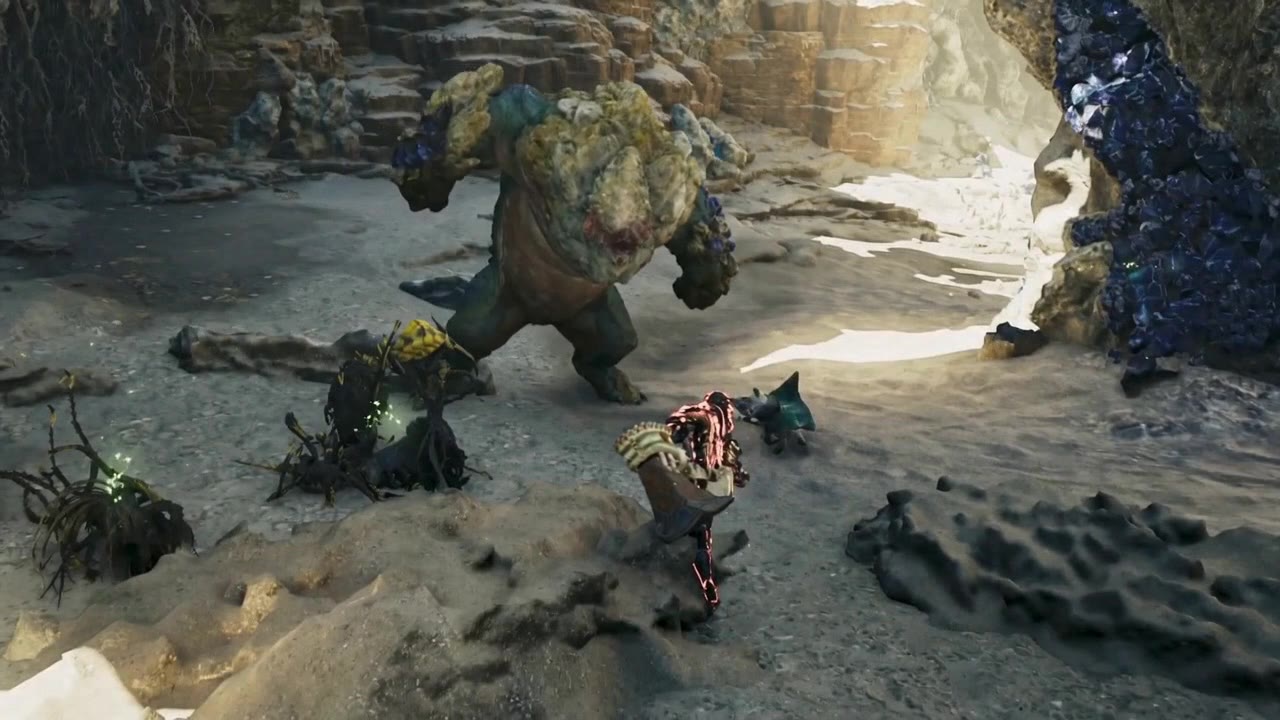} & \includegraphics[width=0.2455\linewidth]{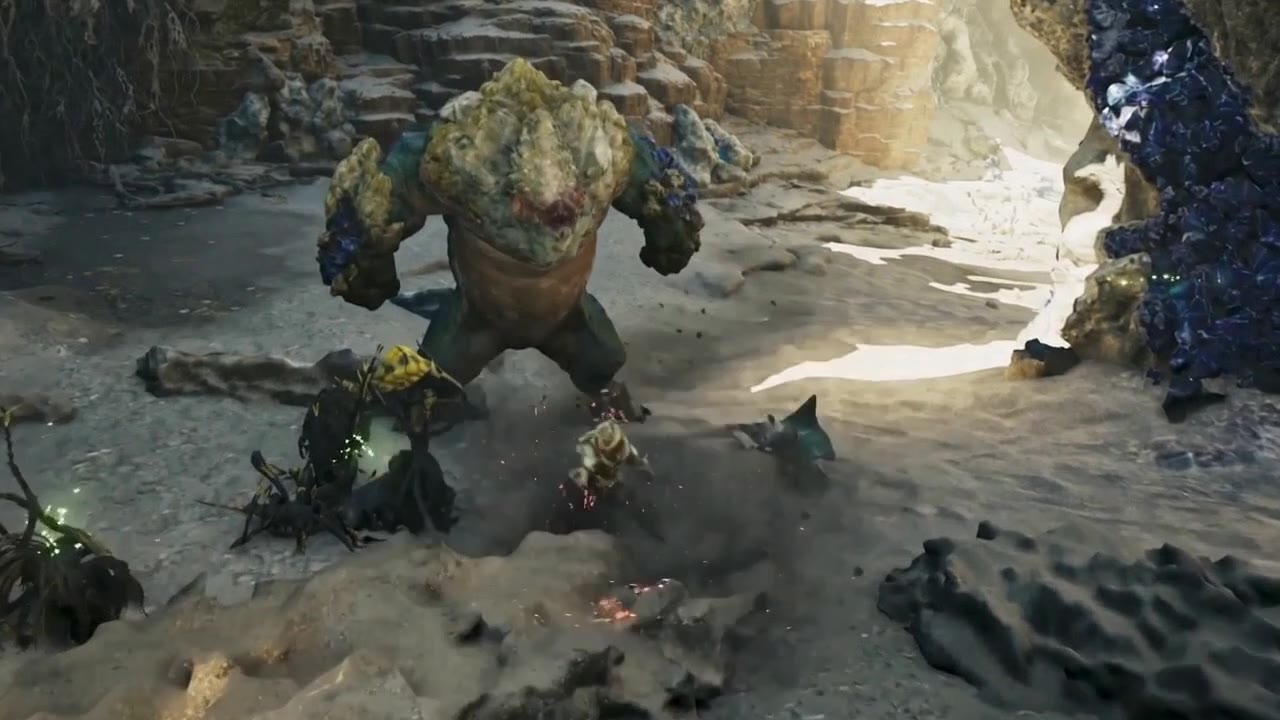} & \includegraphics[width=0.2455\linewidth]{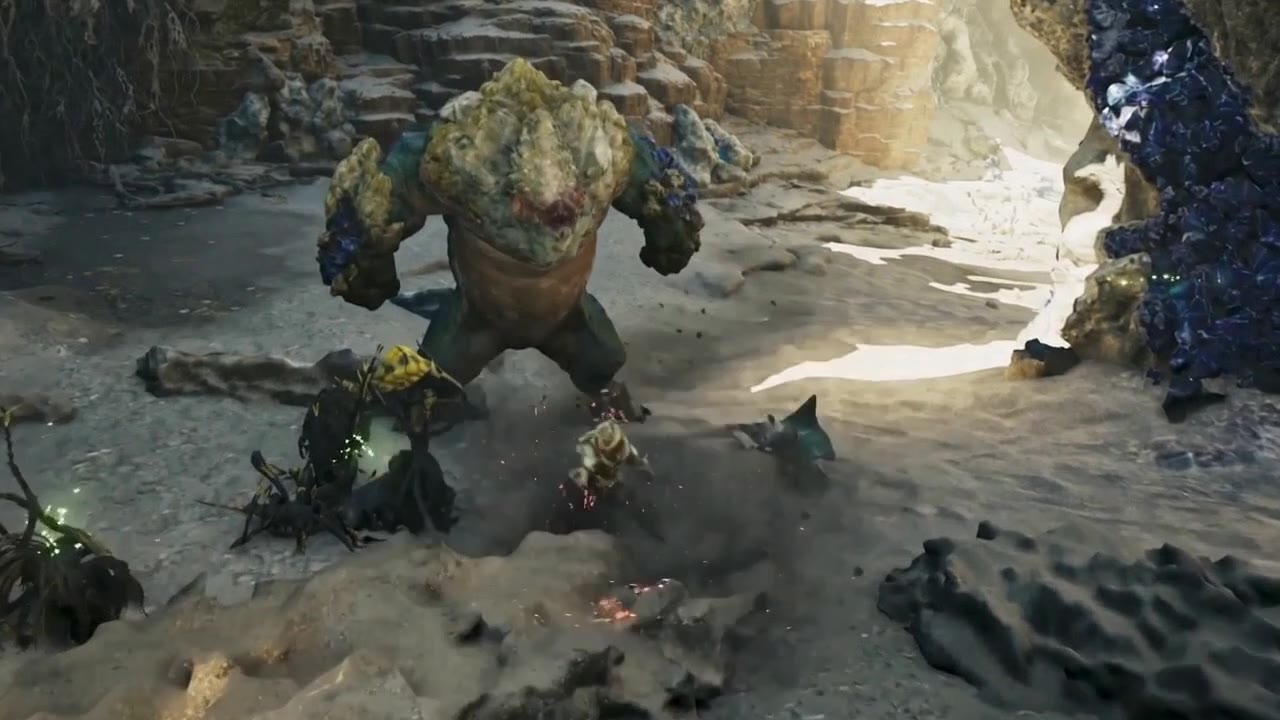} \\[2.5pt]
\multicolumn{4}{@{}l@{}}{\rule{0pt}{2.4ex}{\scriptsize\textbf{Grok Imagine 1.5 (text schedule)}}} \\[0.5pt]
\includegraphics[width=0.2455\linewidth]{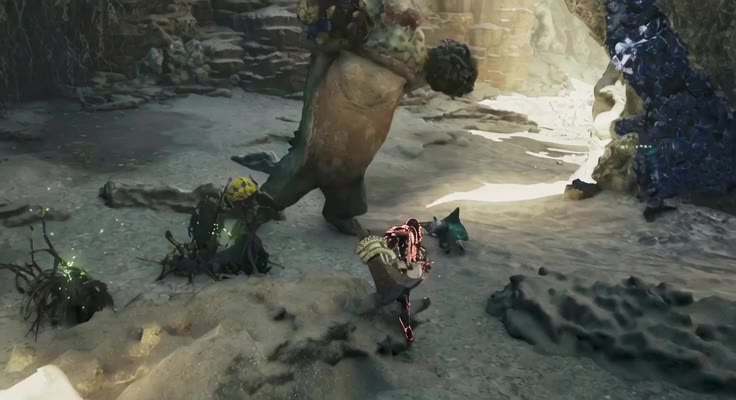} & \includegraphics[width=0.2455\linewidth]{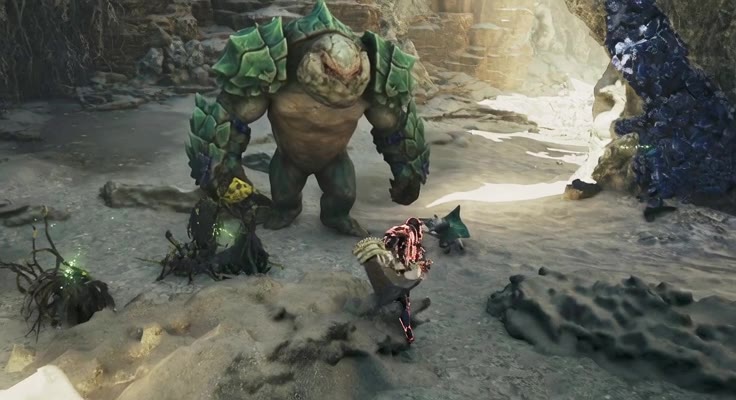} & \includegraphics[width=0.2455\linewidth]{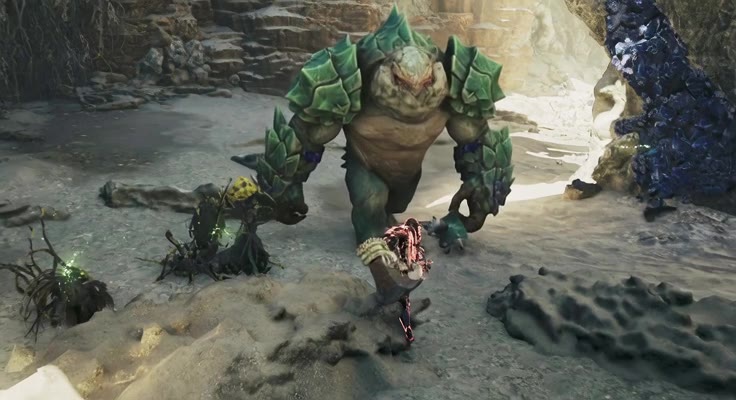} & \includegraphics[width=0.2455\linewidth]{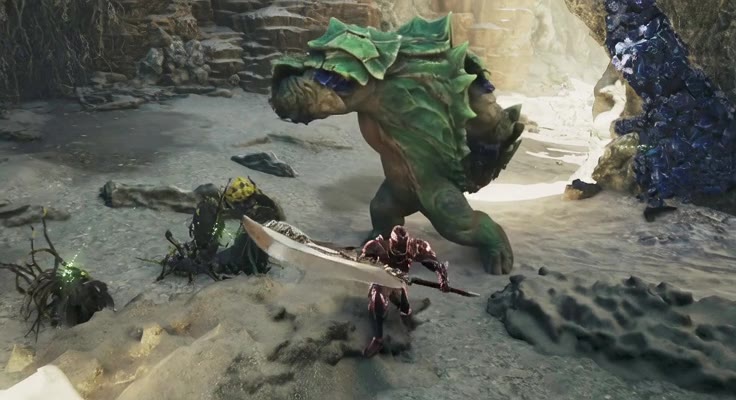} \\
\end{tabular}